\documentclass[lettersize,journal]{IEEEtran}
\usepackage{amsmath,amsfonts}
\usepackage{algorithmic}
\usepackage{algorithm}
\usepackage{array}

\usepackage{textcomp}
\usepackage{stfloats}
\usepackage{url}
\usepackage{verbatim}
\usepackage{graphicx}
\usepackage{cite}

\usepackage{cite}
\usepackage{amsmath,amssymb,amsfonts}
\usepackage{algorithmic}
\usepackage{graphicx}
\usepackage{textcomp}
\usepackage{caption}
\usepackage{subcaption}
\usepackage{algorithm}
\usepackage{booktabs}
\usepackage{tabularx}
\newcolumntype{Y}{>{\centering\arraybackslash}X}
\usepackage{hyperref}

\usepackage{rotating}

\usepackage{color} 
\usepackage{multirow}
 
\begin{document}

\title{RMT-PPAD: Real-time Multi-task Learning for Panoptic Perception in Autonomous Driving}

\author{Jiayuan Wang, Q. M. Jonathan Wu, Katsuya Suto, and Ning Zhang
\thanks{This research was undertaken, in part, thanks to funding from the Canada Research Chairs Program, and in part by the NSERC’s CREATE program on TrustCAV. (Corresponding author: Q. M. Jonathan Wu.)}
\thanks{Jiayuan Wang, Q. M. Jonathan Wu and Ning Zhang are with the Department of Electrical and Computer Engineering, University of Windsor, Windsor, ON N9B 3P4, Canada (e-mails: wang621@uwindsor.ca, jwu@uwindsor.ca and ning.zhang@uwindsor.ca)}
\thanks{Katsuya Suto with the Graduate School of Information Science and Technology at Hokkaido University (e-mail: k.suto@ist.hokudai.ac.jp)}}

\markboth{Journal of \LaTeX\ Class Files,~Vol.~14, No.~8, January~2025}%
{Shell \MakeLowercase{\textit{et al.}}: A Sample Article Using IEEEtran.cls for IEEE Journals}

\maketitle

\begin{abstract}
Autonomous driving systems rely on panoptic driving perception that requires both precision and real-time performance. In this work, we propose RMT-PPAD, a real-time, transformer-based multi-task model that jointly performs object detection, drivable area segmentation, and lane line segmentation. We introduce a lightweight module, a gate control with an adapter to adaptively fuse shared and task-specific features, effectively alleviating negative transfer between tasks. Additionally, we design an adaptive segmentation decoder to learn the weights over multi-scale features automatically during the training stage. This avoids the manual design of task-specific structures for different segmentation tasks. We also identify and resolve the inconsistency between training and testing labels in lane line segmentation. This allows fairer evaluation. Experiments on the BDD100K dataset demonstrate that RMT-PPAD achieves state-of-the-art results with mAP50 of 84.9\% and Recall of 95.4\% for object detection, mIoU of 92.6\% for drivable area segmentation, and IoU of 56.8\% and accuracy of 84.7\% for lane line segmentation. The inference speed reaches 32.6 FPS. Moreover, we introduce real-world scenarios to evaluate RMT-PPAD performance in practice. The results show that RMT-PPAD consistently delivers stable performance. The source codes and pre-trained models are released at \url{https://github.com/JiayuanWang-JW/RMT-PPAD}.
\end{abstract}


\section{Introduction}
\label{sec:introduction}
\begin{figure}[t!]
    \centering
    \includegraphics[width=\linewidth]{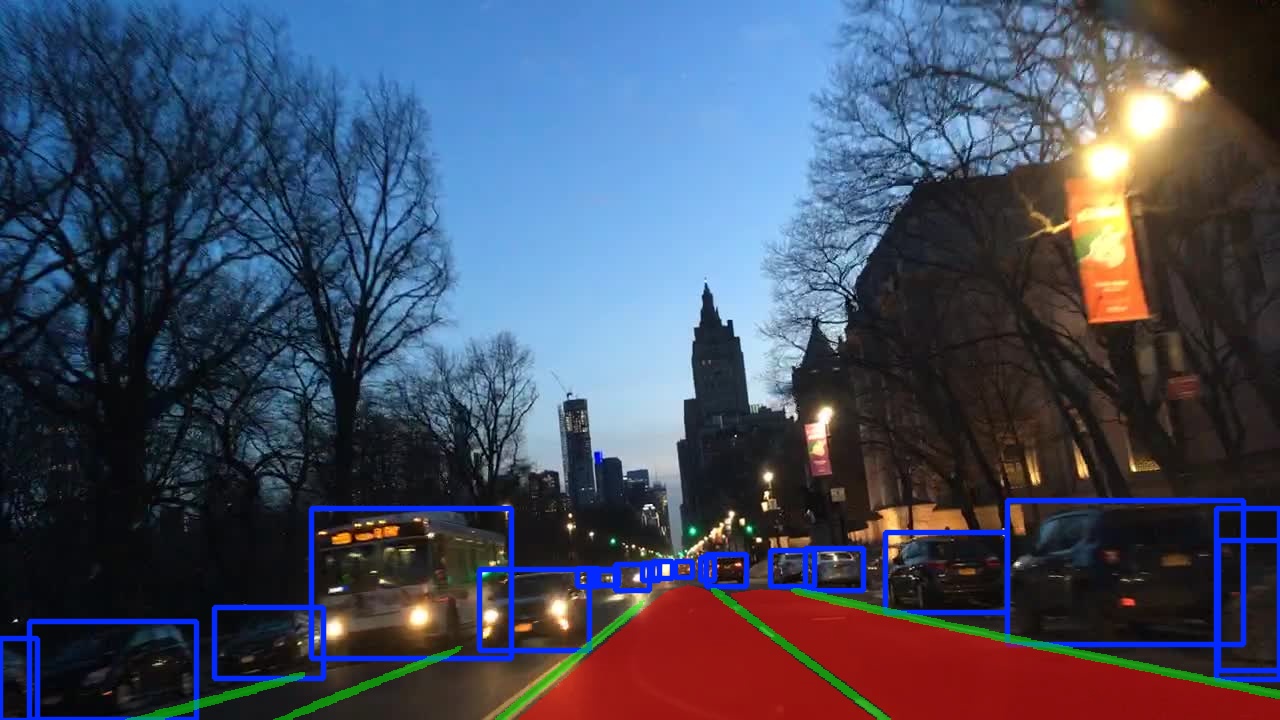}
    \caption{Fundamental tasks in panoptic driving perception: object detection, drivable area and lane line segmentation.}
    \label{fig:intro_fig}
\end{figure}

\IEEEPARstart{T}{he} panoptic driving perception in autonomous driving systems (ADS) needs a complete understanding of the environment in real-time \cite{vandenhende2021multi, yaqoob2019autonomous}. The fundamental tasks in panoptic driving perception are object detection, drivable area segmentation, and lane line segmentation \cite{zhan2024yolopx, wang2024you, wu2022yolop}, as shown in Figure \ref{fig:intro_fig}. Deploying separate on-board models for each task incurs a significant computational cost. It can also hurt real-time performance. Multi-task learning (MTL) is a potential solution to address this challenge by training a single model to handle all tasks jointly. It allows shared feature representations, improving computational efficiency. Additionally, the MTL model can use the relations between tasks to achieve a more coherent scene interpretation \cite{zhang2021loss, ishihara2021multi, wu2022yolop}. For example, road segmentation gives context for object detection. Lane line cues help understand the drivable area. Importantly, the MTL model must also meet real-time inference requirements, with inference speed above 30 FPS to match the camera sampling rate. Because of the limited computing power available on vehicles and the high complexity of the perception tasks \cite{hu2024t, roh2024fast, ouyang2019deep}, achieving high precision on all three tasks with MTL while remaining real-time is very challenging.

\IEEEpubidadjcol

Previous studies demonstrate the promise of MTL in panoptic driving perception, but still show three limitations. The first essential challenge is negative transfer between tasks: the performance on one task decreases when trained together with others due to conflicting features. For example, early methods like YOLOP \cite{wu2022yolop} demonstrate that naively combining detection and segmentation can lead to suboptimal results. The second challenge is that models often require manually designed task-specific structures based on prior knowledge to achieve good performance. For instance, YOLOPX \cite{zhan2024yolopx} employs exclusive head structures for segmentation tasks, which increases design complexity and engineering cost. The third limitation is in the evaluation of lane line segmentation. We identified an inconsistency between training and testing labels in prior works, which leads to incorrect evaluation results. For example, previous related work \cite{zhan2024yolopx, wang2024you, wu2022yolop, vu2022hybridnets, guo2023research, han2022yolopv2} follows \cite{hou2019learning} work, using the 8-pixel width lane lines to train the model and using the 2-pixel width lane lines to evaluate the model performance. This inconsistency leads to better models being evaluated worse because they correctly predict 8-pixel lane lines, but compared to 2-pixel test labels, the extra 6-pixel width is considered the wrong prediction.

To address or alleviate the above limitations, we propose a real-time Transformer-based multi-task model, named RMT-PPAD. For detection parts, RMT-PPAD is built upon RT-DETR \cite{zhao2024detrs}, an object detection model. It avoids heavy post-processing like non-maximum suppression (NMS) and could achieve real-time inference on panoptic driving perception tasks. For the segmentation tasks, we combine multi-scale features from different semantic levels. We propose a learnable weight matrix to adjust the contribution of each scale for each task. This allows the model to learn task-specific feature preferences automatically during the training stage. It avoids manual design of task-specific structures. Additionally, the main contribution of our model is to propose a lightweight gate control with an adapter (GCA) module. Specifically, the adapter extracts task-specific features while retaining shared representations. The gate control module then fuses them adaptively to alleviate negative transfer between tasks. The GCA module is a lightweight design. Therefore, RMT-PPAD achieve outperforming while maintaining real-time performance. Another important contribution of RMT-PPAD is addressing the label inconsistency in lane line segmentation. We demonstrate through experiments and theoretical analysis that using different lane label widths for training and testing is an unfair evaluation. This can wrongly reflect the true performance of models. To address this issue, we align the lane label representation between training and testing. Specifically, we dilate the label widths for the test dataset to the same width as the train dataset. This simple yet effective fix provides a more consistent and fair evaluation for lane line segmentation. 

We through experiments and ablation studies to demonstrate the effectiveness of our proposed model and modules. We evaluate our model and open-source panoptic driving perception MTL models on the BDD100K \cite{yu2020bdd100k} dataset. The results show that RMT-PPAD achieves state-of-the-art (SOTA) performance for all tasks: object detection, drivable area segmentation, and lane line segmentation. While our model remains a real-time inference. Moreover, we also test RMT-PPAD and compare it to YOLOPX on real-world driving scenarios beyond the public dataset. The visualization results show that RMT-PPAD generalizes well and performs reliably in diverse conditions. Our key contributions can be summarized as follows:

\begin{enumerate}
    \item We design a real-time transformer-based multi-task model (RMT-PPAD) without bells and whistles that jointly addresses object detection, drivable area segmentation, and lane line segmentation in a single network.
    \item We propose a lightweight GCA module, which extracts task-specific features, retains shared representations, and adaptively fuses them to alleviate negative transfer between tasks.
    \item We design an adaptive segmentation decoder that learns task-specific weights for multi-scale features automatically. This eliminates the need for manually designed task-specific structures while balancing fine details and global context.
    \item We identify the inconsistency between the lane line label widths used for training and testing in previous works. For a fair and true reflection of the model’s lane line segmentation performance, we propose a simple yet effective method to dilate the test label widths to the same as the train dataset.
    \item We conduct extensive experiments and ablation studies on the BDD100K dataset and real-world driving scenarios to validate the effectiveness of RMT-PPAD, which achieves SOTA performance across all tasks compared to open-source MTL models for panoptic driving perception.
\end{enumerate}

\section{Related Works}

\subsection{Object detection}
Object detection aims to locate and identify objects within an image. Due to the rapid development of deep learning, detection research directions have switched from traditional approaches to deep learning-based methods. Deep learning-based methods can be further divided into anchor-based detectors \cite{girshick2014rich, wang2023yolov7}, anchor-free detectors \cite{hao2025robust, li2024faa}, and Transformer detectors \cite{carion2020end, zhu2020deformable}. 

Anchor-based methods generate dense anchor boxes of various scales and aspect ratios over the image feature map. Then, these anchors are classified as either positive or negative samples based on a specific label assignment strategy. The model subsequently predicts the class and the bounding box offsets for the positive anchors to output the final detection results. However, anchor-based methods meet challenges such as sensitivity to hyperparameters and relying on prior knowledge for aspect ratio. To address these limitations, anchor-free methods were proposed, which allow the model to directly learn object locations and shapes instead of pre-defined anchors. To avoid hand-designed components, such as anchor generation and NMS procedure, a Transformer-based end-to-end method (DETR \cite{carion2020end}) was proposed. It eliminates the need for prior knowledge design by a direct set prediction problem. Specifically, DETR directly predict a fixed number of object queries, each corresponding to a potential object and uses bipartite matching loss to establish a one-to-one correspondence between predicted outputs and ground truth objects. Although DETR achieve outstanding performance, it requires a high computational cost. As a result, they are difficult to deploy in real-time systems. This limitation restricts their use in time-sensitive applications such as autonomous driving. Therefore, RT-DETR \cite{zhao2024detrs} was proposed to address the challenge. It achieves efficient and real-time object detection by integrating an efficient hybrid encoder. In this work, we are based on RT-DETR to build RMT-PPAD because it achieves a good balance between accuracy, inference time, and computational complexity for the detection task. 

\begin{figure*}[ht]
    \centering
    \includegraphics[width=\linewidth]{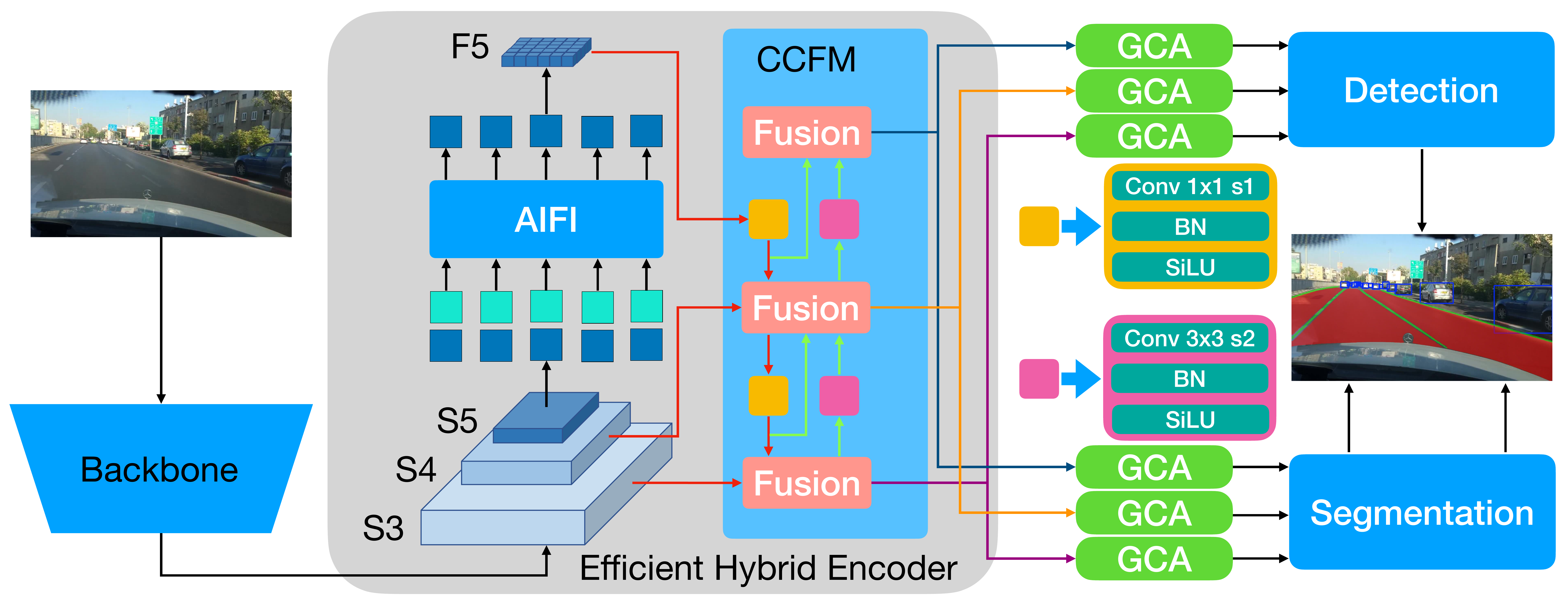}
    \caption{Overview of RMT-PPAD structure.}
    \label{fig:sturcture}
\end{figure*}

\subsection{Semantic segmentation}
Unlike object detection, semantic segmentation is a pixel-level operation. It requires classifying each pixel, including the background. To achieve accurate segmentation, high-resolution multi-scale features are essential. They capture both fine-grained spatial details and high-level semantic information. The semantic segmentation methods are mainly divided into region-based \cite{caesar2016region, girshick2015region, he2017std2p}, fully convolutional network-based, and transformer-based \cite{cheng2021per, cheng2022masked} methods. 

Region-based semantic segmentation first proposes candidate regions and encodes them with features. Then, each region is classified independently, and the predicted label is propagated to all pixels within the region. However, this approach has a significant limitation in real-time inference applications because it has a multi-stage pipeline instead of an end-to-end pipeline. To address this limitation, fully convolutional networks \cite{long2015fully} (FCN) were proposed. It directly predicts classification for each pixel. FCN can perform efficient pixel-level inference and end-to-end training by replacing fully connected layers with convolutional layers. However, its limited receptive field hinders global context modelling, often producing coarse segmentation. To address this limitation, DeepLab \cite{chen2017deeplab} and PSPNet \cite{zhao2017pyramid} further improved FCN-based models by incorporating multi-scale context, atrous convolution, and pyramid pooling modules to better handle objects at different scales and improve boundary precision. Recently, Transformer-based methods were proposed, such as MaskFormer \cite{cheng2021per} and Mask2Former \cite{cheng2022masked}. They formulate segmentation as a set prediction problem, where each output consists of a class and a soft mask representing a specific semantic region. They use a Transformer decoder with a set of learnable queries to directly predict a fixed number of semantic masks instead of performing per-pixel classification. This enables end-to-end segmentation with better global context modelling and flexible region assignment. Similar to Transformer-based detection models, they struggle to achieve real-time performance. 

\subsection{Multi-task learning}
MTL aims to perform multiple tasks simultaneously in a single model by sharing the backbone and separate heads. Compared to deploying separate models for each task, MTL can reduce deployment costs and accelerate inference. This advantage makes MTL well-suitable for applications that require real-time performance, such as ADS. Currently, many recent works have adopted MTL in the panoptic driving perception, which usually includes tasks such as vehicle detection, drivable area segmentation, and lane line segmentation. For example, YOLOP \cite{wu2022yolop} YOLOP introduces a simple yet effective architecture to achieve outstanding performance on BDD100K. Nevertheless, there is a significant issue with YOLOP: negative transfer. Their ablation study shows except for Recall in object detection, all of the other metrics are decreased compared to single-task learning. Although following YOLOP, there are several works were proposed \cite{zhan2024yolopx, vu2022hybridnets, wang2024you, fang2024yolomh}, none were aware of the negative transfer issue. However, negative transfer is a classical challenge in MTL. It is critical to pay attention to negative transfer to avoid performance decrease to undermines the advantage of joint learning.

\section{Proposed Methodology}

\subsection{Model Overview}

We base on RT-DETR \cite{zhao2024detrs} to design RMT-PPAD, which consists of a backbone, an efficient hybrid encoder, six GCAs, and two task-specific decoders. The overview of our proposed model is illustrated in Fig. \ref{fig:sturcture}. Specifically, we follow RT-DETR to use high-performance GPU network V2 (HGNetV2 \footnote{\url{https://github.com/PaddlePaddle/PaddleDetection/blob/develop/ppdet/modeling/backbones/hgnet_v2.py}}) with L scale as the backbone to extract the features. Then, we feed the last three stages' features (S3, S4, S5) from the backbone into the efficient hybrid encoder. An efficient hybrid encoder consists of attention-based intra-scale feature interaction (AIFI) and the CNN-based cross-scale feature-fusion module (CCFM) to rapidly process the multi-scale features. Subsequently, the multi-scale feature maps from the CCFM feed into the three GCAs respectively to alleviate the negative transfer challenge (\textit{cf.} Sec. \ref{sec:GCA}). Finally, the fused multi-scale features are fed into detection and segmentation decoders generate the outputs for object detection, drivable area segmentation, and lane line segmentation (\textit{cf.} Sec. \ref{sec:decoders}). 

\subsection{Gate control with an adapter}
\label{sec:GCA}

\begin{figure}[ht]
    \centering
    \includegraphics[width=\linewidth]{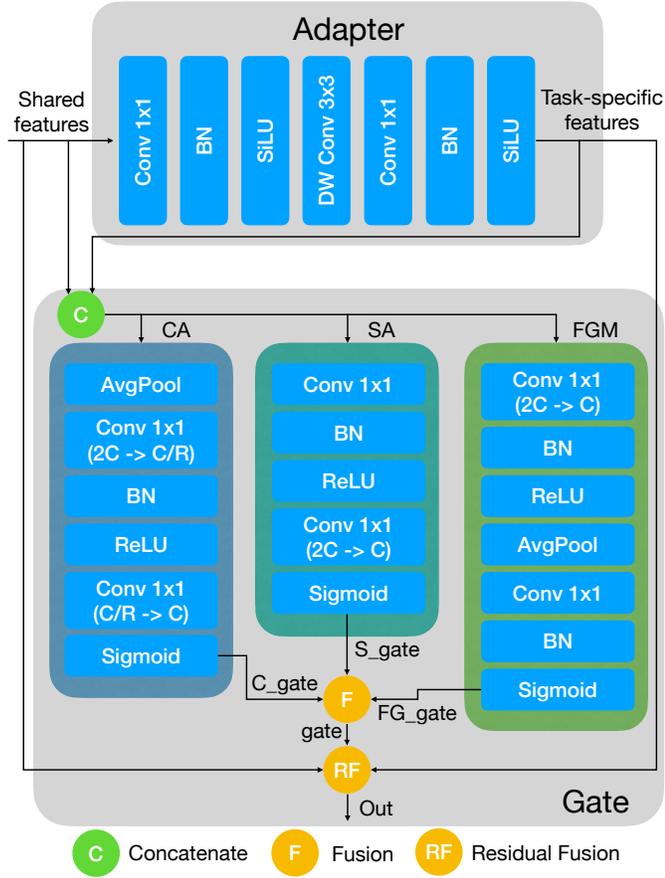}
    \caption{Overview of GCA structure. DW Conv denotes depthwise convolution. C is the number of channels, and R is the channel reduction ratio in attention modules.}
    \label{fig:GCA}
\end{figure}
To alleviate the negative transfer challenges, we propose the GCA module. It could capture task-specific features from shared representations and adaptively fuse them with shared features. This module allows each task to emphasize relevant shared features while suppressing irrelevant features, thereby reducing gradient conflicts and alleviating negative transfer. The GCA structure is illustrated in Fig. \ref{fig:GCA}. Specifically, it consists of a task adapter and a dynamic gate. The multi-scale features (S3, S4, S5) outputs from the efficient hybrid encoder feed into GCA separately. Thanks to the lightweight design of GCA, which is built on simple convolutional layers with intermediate channel reduction to minimize computing overhead. Therefore, we can apply separate GCA at each scale to effectively capture fine-grained task-specific features while maintaining real-time performance. The adapter efficiently captures task-specific representation for each scale feature without disrupting the shared features. It consists of a 1x1 convolution for cross-channel transformation, followed by a depthwise separable convolution. Additionally, batch normalization (BN) and sigmoid linear unit (SiLU) are used after each block. The output task-specific features are concatenated with the shared features along the channel dimension to form a combined representation for subsequent gating.

For the gate part, it primarily includes three modules: channel attention (CA), spatial attention (SA), and fusion gate module (FGM), their outputs are $C\_gate$, $S\_gate$, and $FG\_gate$, respectively. Specifically, we adopt a squeeze-and-excitation \cite{hu2018squeeze} (SE) style CA, which learn what channels to emphasize when fusing the two streams. In contrast, SA is responsible for learning which spatial locations (pixels) within each channel are most important. The last model FGM is responsible for learns a per-channel gating map that dynamically balances global channel attention and local spatial attention. Then, we use $FG\_gate$ to fuse the $C\_gate$ and $S\_gate$ to obtain the attention map ($gate$). The formula is shown as follows:
\begin{equation}
gate = FG\_gate \cdot C\_gate + (1 - FG\_gate) \cdot S\_gate
\end{equation}
Additionally, we use a clipping function to constrain the gate within the range [0.05, 0.95]. This is avoiding either shared or task-specific features from being fully suppressed or fully dominant.

Finally, we get the final output through a residual interpolation to fuse the shared features ($shared$) and task-specific features ($task$). The formula is shown as follows:
\begin{equation}
out = shared + gate \cdot (task - shared)
\end{equation}

\subsection{Decoders}
\label{sec:decoders}
The decoder processes the feature maps from the encoder to produce task-specific predictions. This includes predicting object classes, their corresponding bounding boxes, and masks. In our work, we implement this using two separate decoders: the detection decoder and the segmentation decoder. We will introduce them one by one. 

\begin{figure}[ht]
    \centering
    \includegraphics[width=\linewidth]{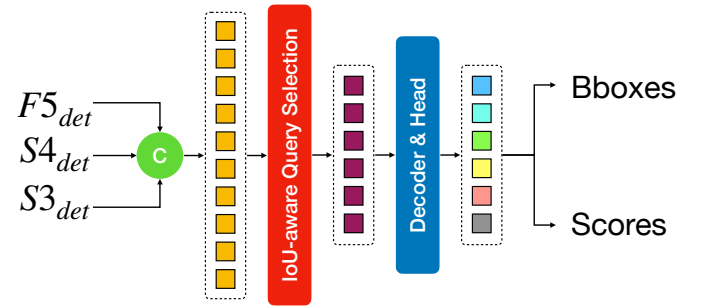}
    \caption{Structure of the detection decoder. Yellow boxes are multi-scale feature tokens, purple boxes are the selected object queries, and the colored boxes illustrate the predicted classification scores and bounding box for each object query.}
    \label{fig:sdetection}
\end{figure}

\emph{Detection decoder:}
Due to RT-DETR \cite{zhao2024detrs} decoder having an outstanding balance between speed and accuracy performance, we follow its design for our object detection decoder. This decoder reduces inference time and avoids the operation of NMS in post-processing. The structure of detection is shown in Fig. \ref{fig:sdetection}. Specifically, its inputs are three different scale feature maps ($S3_{det}$, $S4_{det}$, and $F5_{det}$), which are from three GCAs in the detection branch. First, flatten each feature map and concatenate them into one sequence. Then, we adopt IoU-aware query selection to select a fixed number of encoder features as initial object queries for the decoder. Finally, the decoder with auxiliary prediction heads iteratively optimizes object queries to generate bounding boxes (Bboxes) and confidence scores (Scores). 

\begin{figure}[ht]
    \centering
    \includegraphics[width=\linewidth]{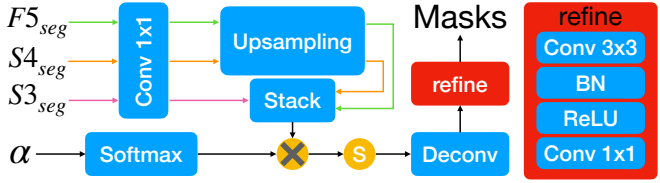}
    \caption{Structure of Segmentation. $\alpha$ is a learnable weight tensor. }
    \label{fig:sseg}
\end{figure}

\begin{figure*}[hb]
    \centering
    \begin{subfigure}[t]{0.23\textwidth}
        \centering
        \includegraphics[width=\textwidth]{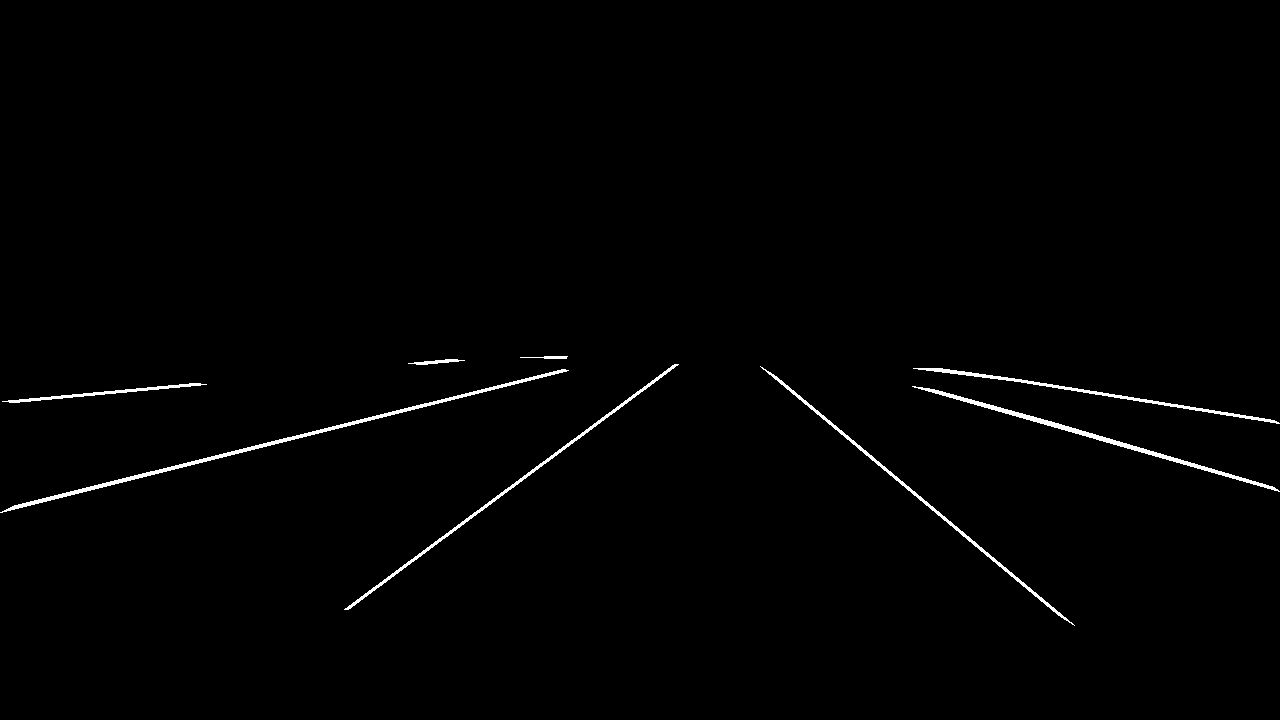} 
        \caption{2-pixel GT}
        \label{fig:sub_lane_a}
    \end{subfigure}
    \hfill
    \begin{subfigure}[t]{0.23\textwidth}
        \centering
        \includegraphics[width=\textwidth]{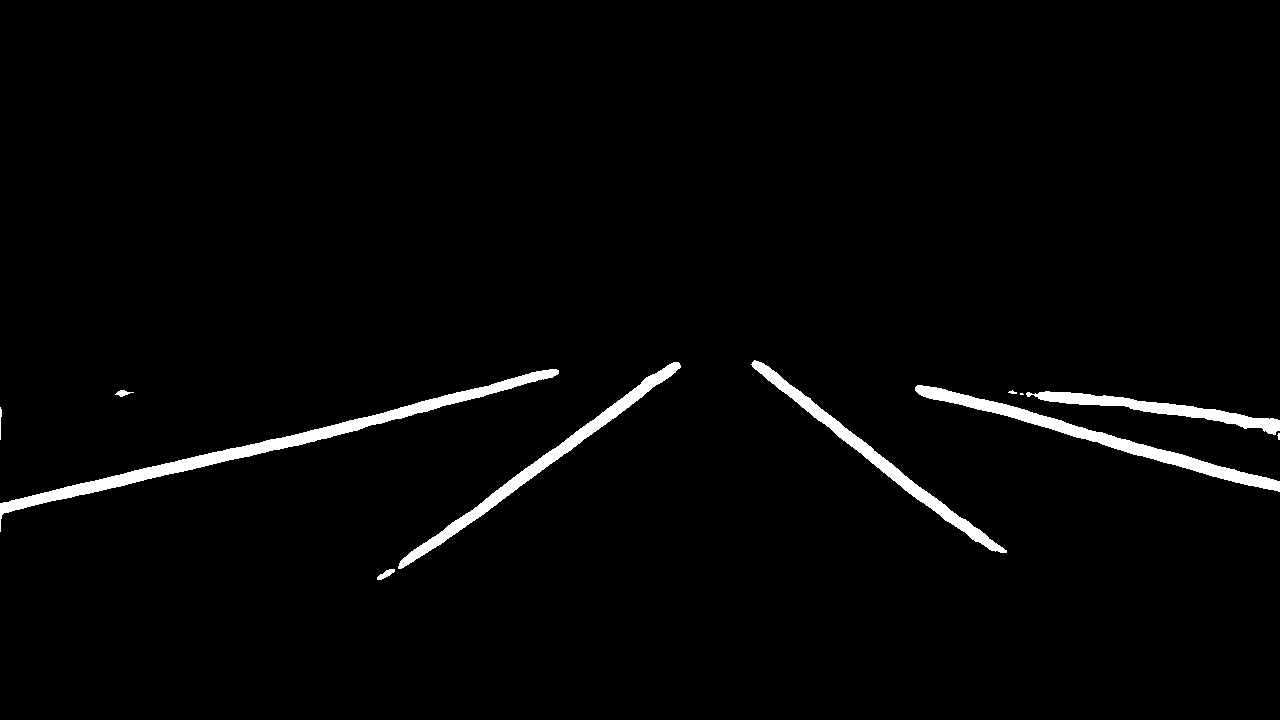} 
        \caption{A-YOLOM}
        \label{fig:sub_lane_b}
    \end{subfigure}
    \hfill
    \begin{subfigure}[t]{0.23\textwidth}
        \centering
        \includegraphics[width=\textwidth]{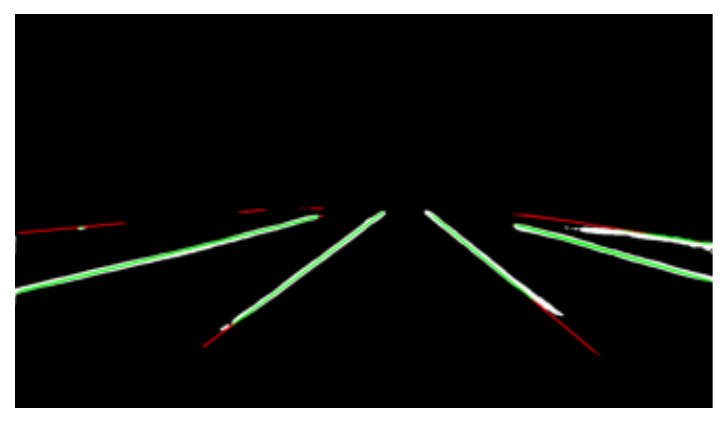} 
        \caption{A-YOLOM vs 2-pixel GT \\ IoU=0.2612, ACC=0.7191}
        \label{fig:sub_lane_c}
    \end{subfigure}
    \hfill
    \begin{subfigure}[t]{0.23\textwidth}
        \centering
        \includegraphics[width=\textwidth]{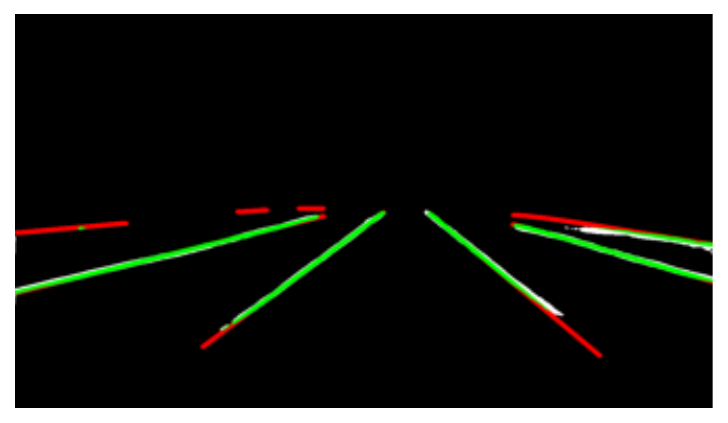} 
        \caption{A-YOLOM vs 8-pixel GT \\ IoU=0.5058, ACC=0.6457}
        \label{fig:sub_lane_d}
    \end{subfigure}
    
    \vskip\baselineskip 

    \begin{subfigure}[t]{0.23\textwidth}
        \centering
        \includegraphics[width=\textwidth]{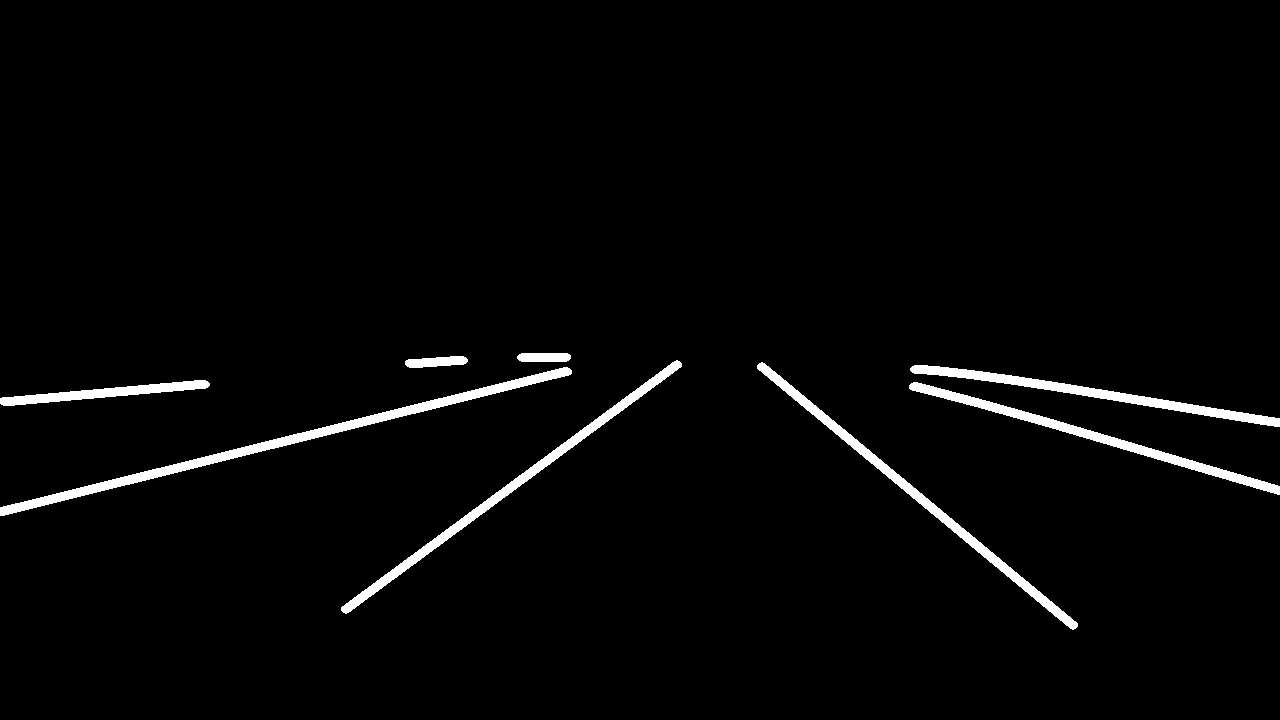} 
        \caption{8-pixel GT}
        \label{fig:sub_lane_e}
    \end{subfigure}
    \hfill
    \begin{subfigure}[t]{0.23\textwidth}
        \centering
        \includegraphics[width=\textwidth]{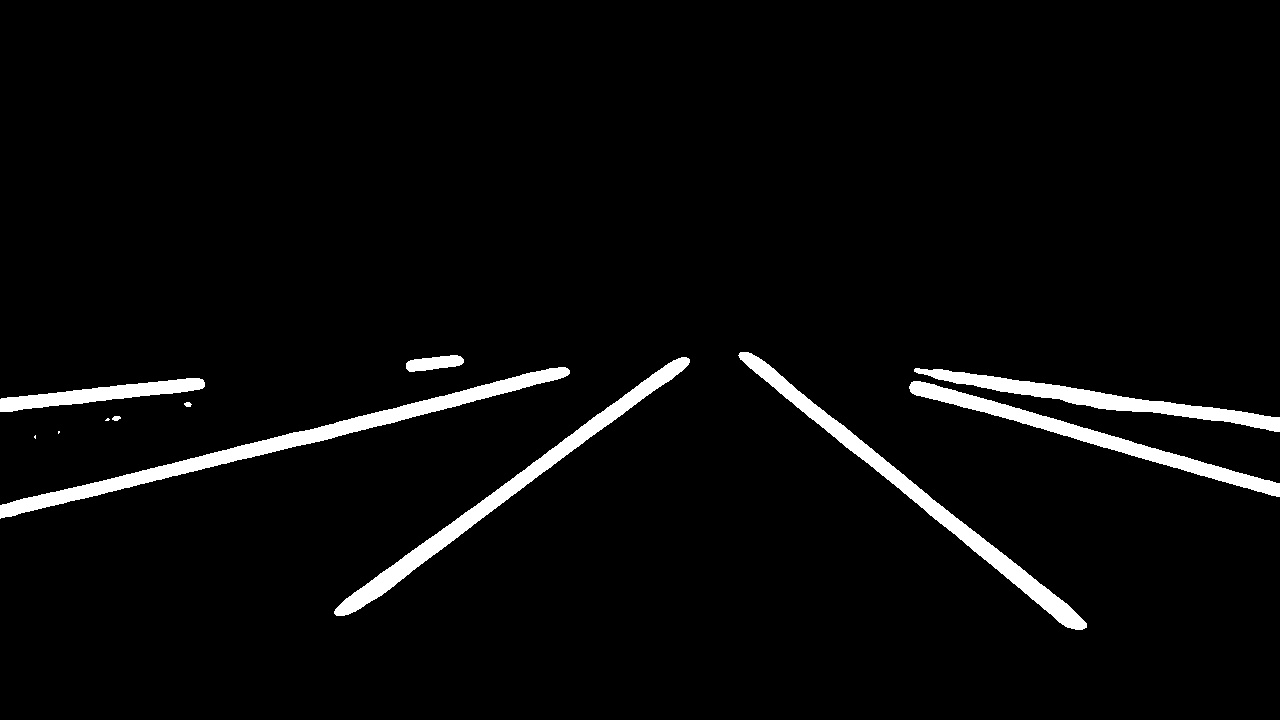} 
        \caption{YOLOPX}
        \label{fig:sub_lane_f}
    \end{subfigure}
    \hfill
    \begin{subfigure}[t]{0.23\textwidth}
        \centering
        \includegraphics[width=\textwidth]{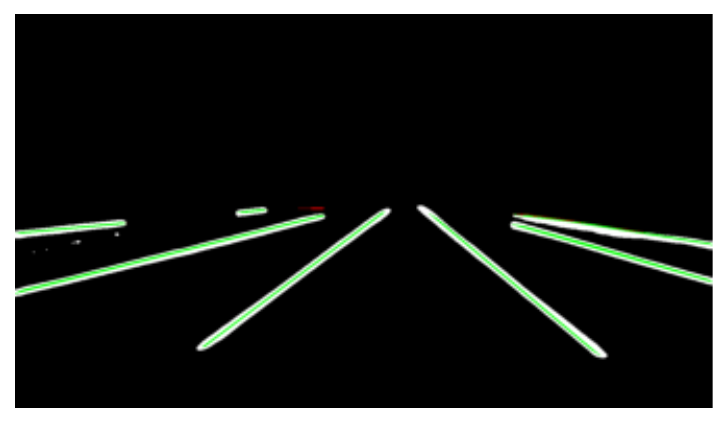} 
        \caption{YOLOPX vs 2-pixel GT \\ IoU=0.2339, ACC=0.9665}
        \label{fig:sub_lane_g}
    \end{subfigure}
    \hfill
    \begin{subfigure}[t]{0.23\textwidth}
        \centering
        \includegraphics[width=\textwidth]{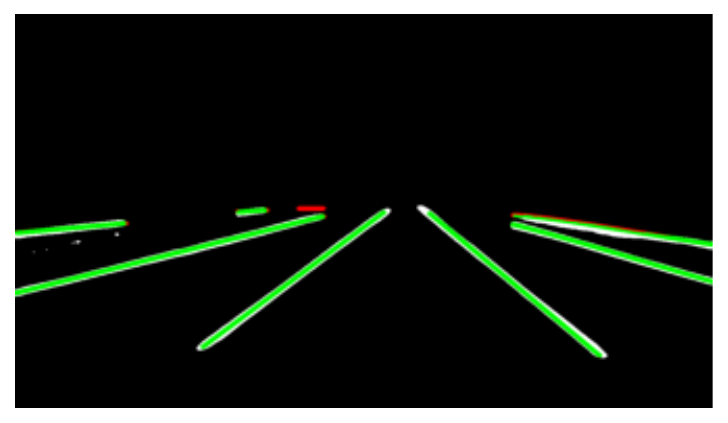} 
        \caption{YOLOPX vs 8-pixel GT \\ IoU=0.5851, ACC=0.9338}
        \label{fig:sub_lane_h}
    \end{subfigure}
    
    \caption{Lane line test label compares explanation. (a) is ground truth (GT) for 2-pixel, which is the previous work used label for the lane line test. (e) is our proposed dilating to 8-pixel GT. (b) and (f) are predictions from A-YOLOM(n) and YOLOPX model, separately. (c) and (g) are comparisons between 8-pixel GT (red) and prediction (white), where green colour means overlap area. (d) and (h) are comparisons between 8-pixel GT and prediction.}
    \label{fig:lane_proof}
\end{figure*}

\emph{Segmentation decoder:}
We adopt a unified segmentation decoder for two tasks. This design is different from previous similar works, such as A-YOLOM \cite{wang2024you}, YOLOPX \cite{zhan2024yolopx}, and HybridNet \cite{vu2022hybridnets}. We perform all the segmentation tasks in a unified architecture. This ensures outstanding performance while making the model as lightweight as possible and achieving real-time inference. Additionally, in previous work, it usually designs different structures for different segmentation tasks because they rely on different scale feature maps. This requires prior knowledge and extensive experiments for tuning. In our work, we design an adaptive structure to automatically learn the weights for different-scale feature maps during the training stage. Our proposed structure of segmentation is illustrated in Fig. \ref{fig:sseg}. Specifically, the input are multi-scale feature maps ($S3_{seg}$, $S4_{seg}$, and $F5_{seg}$) from segmentation branch three GCAs. We first feed each feature map into a 1x1 convolutional layer to project all the features into the same number of channels for subsequent steps. Then we upsample $S4_{seg}$ and $F5_{seg}$ match to the resolution of $S3_{seg}$ through bilinear interpolation. We then stack the feature maps along a new dimension, producing a tensor that aggregates multi-scale information for subsequent weighting. Meanwhile, we adopt a learnable 2×3 tensor $\alpha$, which is normalized via softmax, to get weights for each scale’s feature map in each task. The value of $\alpha$ is adjusted automatically by training. Next, we broadcast $\alpha$ to the shape of the stacked feature tensor and perform element-wise multiplication, thereby weighting each scale’s feature map. Then, sum over the dimension of the stacked tensor to produce a task-specific feature map. However, this task-specific feature map is $S3$ size, so we use deconvolutional layers to recover the feature maps to the original input image size. Finally, feed the upsampled feature map a refinement head to produce one individual mask for each segmentation task.

\subsection{Lane line label and evaluation metric}
\label{ll_lable_metrics}

In this subsection, we will demonstrate that the lane line test label and metrics used in previous studies have certain significant issues and propose more reasonable labels for lane line testing. 

In \cite{zhan2024yolopx}, the authors proposed two opinions on lane line evaluation metrics: 1. The evaluation metric intersection over union (IoU) is generally low. 2. Pixel accuracy reflects lane detection performance more accurately. Following these opinions, we have an in-depth exploration to identify the potential reasons and assess their rationality. We through an example combining visual and numerical results to demonstrate two issues that we found: 

{ 
\renewcommand{\arraystretch}{2.0}  
\setlength{\tabcolsep}{12pt}       

\begin{table*}[ht]
    \centering
    \caption{Confusion matrix of Fig. \ref{fig:sub_lane_c}, 
    \label{TABLE:Confusionmatrix}
    \ref{fig:sub_lane_d}, \ref{fig:sub_lane_g}, and \ref{fig:sub_lane_h}.}
    \begin{subtable}[t]{0.19\textwidth}
        \centering
        \begin{tabular}{|c|c|}
            \hline
            TN & FP\\ \hline
            FN & TP\\ \hline
        \end{tabular}
        \vspace{2mm} 
        \caption{Confusion matrix}
    \end{subtable}%
    \hfill
    \begin{subtable}[t]{0.19\textwidth}
        \centering
        \begin{tabular}{|c|c|}
            \hline
            898453 & 14738\\ \hline
            2362   & 6047\\ \hline
        \end{tabular}
        \vspace{2mm}
        \caption{Fig. \ref{fig:sub_lane_c}}
        \label{subtable:cm_b}
    \end{subtable}%
    \hfill
    \begin{subtable}[t]{0.19\textwidth}
        \centering
        \begin{tabular}{|c|c|}
            \hline
            892833 & 6235\\ \hline
            7982   & 14550\\ \hline
        \end{tabular}
        \vspace{2mm}
        \caption{Fig. \ref{fig:sub_lane_d}}
        \label{subtable:cm_c}
    \end{subtable}
    \hfill
    \begin{subtable}[t]{0.19\textwidth}
        \centering
        \begin{tabular}{|c|c|}
            \hline
            886849 & 26342\\ \hline
            282    & 8127\\ \hline
        \end{tabular}
        \vspace{2mm}
        \caption{Fig. \ref{fig:sub_lane_g}}
        \label{subtable:cm_d}
    \end{subtable}%
    \hfill
    \begin{subtable}[t]{0.19\textwidth}
        \centering
        \begin{tabular}{|c|c|}
            \hline
            885640 & 13428\\ \hline
            1491   & 21041\\ \hline
        \end{tabular}
        \vspace{2mm}
        \caption{Fig. \ref{fig:sub_lane_h}}
        \label{subtable:cm_e}
    \end{subtable}%
\end{table*}
} 

1) The lane line labels used in previous studies for testing are unfair. 

To show this issue, we adopted a sample named b9e91422-944c7a2f from the BDD100K \cite{yu2020bdd100k} dataset. Compare Fig. \ref{fig:sub_lane_b} and \ref{fig:sub_lane_f} to \ref{fig:sub_lane_a}, YOLOPX is much better and more comprehensive than A-YOLOM via visualization evaluation. However, the quantitative result of IoU is the converse. Comparing results in Fig. \ref{fig:sub_lane_c} and \ref{fig:sub_lane_g}, the IoU result demonstrates that A-YOLOM is better than YOLOPX. This conflict results because the test label uses 2-pixel widths for the lane line. However, the training label uses 8-pixel widths \cite{hou2019learning}. Therefore, the predicted mask is more accurate, the false positive ($FP$) is higher, which will negatively impact the IoU, further wrong to evaluate the model performance. Next, we will use formulas and confusion matrix to prove it.

The formula for IoU metrics is as follows:
\begin{equation}
\text{IoU} = \frac{TP}{TP + FN + FP}
\end{equation}
where $TP$ corresponds to green areas in Fig. \ref{fig:sub_lane_c}, \ref{fig:sub_lane_d}, \ref{fig:sub_lane_g}, and \ref{fig:sub_lane_h}, which means the true foreground and correctly predicted as foreground. $FN$ corresponds to red areas, which means true foreground but are predicted as background (missed detections). $FP$ corresponds to white areas, which means true background but are predicted as foreground (false detections). The confusion matrix of Fig. \ref{fig:sub_lane_c}, \ref{fig:sub_lane_d}, \ref{fig:sub_lane_g}, and \ref{fig:sub_lane_h} as shown in Table \ref{TABLE:Confusionmatrix}, where $TN$ is true negative means true background and correctly predicted as background. 

\begin{figure}[htb]
    \centering
    \includegraphics[width=0.5\linewidth]{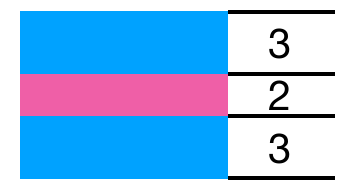}
    \caption{Illustration of lane lines. The pink area is a 2-pixel lane line. And pink plus blue area is a 8-pixel lane line.}
    \label{fig:lane_line}
\end{figure}

Compare Table \ref{subtable:cm_b} and Table \ref{subtable:cm_d}, The IoU of Table \ref{subtable:cm_d} is worse than Table \ref{subtable:cm_b}. The main issue is $FP$ in Table \ref{subtable:cm_d} is much higher than Table \ref{subtable:cm_b}. High $FP$ means the white area is too much. The reason is predicted lane has 8-pixel width and the ground truth only has 2-pixel width. The illustration in Fig. \ref{fig:lane_line}, where pink area is $TP$, and blue area is $FP$. Therefore, the model predicts a more accurate lane line ($TP$) will introduce more $FP$, and the ideal situation increase rate is TP:FP $=$ 1:3. 

We check the label used by previous works (2-pixel) and found that it is not strictly 2 pixels for all of the lane lines, some are 3 pixels, and slashes may include square root values. After dilating operation, some lane lines expand to 8-pixel and some of them expand to 9-pixel, depending on the original label size and slashes may include square root values as well. This phenomenon also exists in training samples. Specifically, we apply morphological dilation using a 7×7 elliptical structuring element. A lane line that is horizontal or vertical extends precisely 3 pixels in both vertical line directions. However, for slash, it will introduce systematic error because the elliptical structuring element is an approximate circle, not a standard circle. However, we believe this tiny systematic error is acceptable. Comparing Fig. \ref{fig:sub_lane_d} and Fig. \ref{fig:sub_lane_h}, the IoU correctly demonstrate the YOLOPX is better than A-YOLOM and this result corresponds to visual evaluation. Table \ref{subtable:cm_c} and \ref{subtable:cm_e} shows the corresponding confusion matrix, which indicates that after dilating the label predicted mask closer to ground truth, more $TP$ and less $FP$ compared to previous work labels in Table \ref{subtable:cm_b} and \ref{subtable:cm_d}. These results demonstrate that the after-dilating label matches the training label. Therefore, our proposed dilated label more accurately and fairly reflects the model's performance.

2) Pixel accuracy metrics do not comprehensively reflect the true lane line performance of the model. 

In previous works \cite{wu2022yolop, zhan2024yolopx, han2022yolopv2, fang2024yolomh}, all of them use pixel accuracy (ACC) as the evaluation metric for lane line segmentation. However, they essentially use Recall to calculate the ACC for the lane line task. Reference YOLOP \cite{wu2022yolop} lane line accuracy formula \footnote{\url{https://github.com/hustvl/YOLOP/blob/main/lib/core/evaluate.py\#L198}} is as follows:
\begin{equation}
\text{LineAccuracy} = \frac{TP}{TP + FN}
\end{equation}
This ACC formula does not penalize over-segmentation. A model can label many extra background pixels as lane lines and still score high ACC because $FP$ is not included in the formula.

{ 
\renewcommand{\arraystretch}{2.0}  
\setlength{\tabcolsep}{12pt}      

\begin{table}[ht]
    \centering
    \caption{Confusion matrix for 8-pixel GT with 2-pixel GT.} 
    \label{TABLE:Confusionmatrix_8_2}
    \begin{tabular}{|c|c|}
        \hline
        899068 & 14123 \\
        \hline
        0   & 8409  \\
        \hline
    \end{tabular}
\end{table}
} 

We experiment with the 8-pixel GT as the predicted mask and the 2-pixel GT as the label. The confusion matrix in Table \ref{TABLE:Confusionmatrix_8_2}. The result of ACC is 1.0, which means as long as the predicted mask includes the lane line, the ACC will be high because $FN$ is lower or 0. Therefore, focusing on lane line pixel accuracy alone \cite{zhan2024yolopx} can not accurately and reliably reflect the model's true performance. It should be combined with other metrics to evaluate, such as IoU. 

\subsection{Loss functions}
RMT-PPAD adopts an end-to-end structure training strategy. Therefore, the objective included multiple loss functions for different tasks. Specifically, it is composed of one detection and two segmentation loss functions. The formula is shown as follows: 
\begin{equation}
\mathcal{L}=\mathcal{L}_{\text {det }}+\mathcal{L}_{\text {segda }}+\mathcal{L}_{\text {segll }}
\end{equation}
where $\mathcal{L}_{\text {det}}$ for object detection task, $\mathcal{L}_{\text {segda}}$ for drivable area segmentation task, and $\mathcal{L}_{\text {segll}}$ for lane line segmentation task. 

\subsubsection{Detection task} RMT-PPAD follow RT-DETR \cite{zhao2024detrs} use the standard DETR \cite{carion2020end} bipartite matching loss, and add a denoising branch to accelerate convergence and stabilize queries. 

Denote a fixed-size set of $N$ predictions by $\{(\hat b_i,\hat p_i)\}_{i=1}^N,$
where $\hat b_i\in\mathbb{R}^4$ are bounding‐box coordinates and $\hat p_i\in\mathbb{R}^{C}$ are class logits, where $C$ is the number of the classification in detection. The $M$ ground‐truth objects are
$\{(b_j,c_j)\}_{j=1}^M,$ $\quad b_j\in\mathbb{R}^4,$ $\;c_j\in\{1,\dots,C\}$. The overall $\mathcal{L}_{\text {det}}$ is combined by two parts:
\begin{equation}
\mathcal{L}_{\text {det}}
= \mathcal{L}_{\text {DETR}} + \mathcal{L}_{\text {dn}}
\end{equation}
where $\mathcal{L}_{\text {DETR}}$ follow DETR \cite{carion2020end} work, which include two parts:
\begin{equation}
\mathcal{L}_{\text {DETR}}
= \mathcal{L}_{\text {core}} + \mathcal{L}_{\text {aux}}
\end{equation}
Specifically, it first uses the Hungarian algorithm to find the optimal one-to-one matching $\sigma^*$ between predictions and ground-truth targets. Then, it minimizes a matching cost combining classification probability and bounding-box similarity. After matching, it computes a loss as $\mathcal{L}_{\text {core}}$: 

\begin{equation}
L_{\mathrm{core}}
= L_{\mathrm{cls}}^{\mathrm{match}}
+ L_{\mathrm{bbox}}
+ L_{\mathrm{cls}}^{\mathrm{unmatch}}
\end{equation}

\begin{equation}
L_{\mathrm{cls}}^{\mathrm{match}}
= \frac{\alpha}{M}
\sum_{j=1}^{M}
L_{\mathrm{cls}}\bigl(\hat p_{\sigma^*(j)},\,c_j\bigr)
\end{equation}

\begin{equation}
L_{\mathrm{bbox}}
= \frac{\beta}{M}
\sum_{j=1}^{M}
\bigl\lVert b_j - \hat b_{\sigma^*(j)}\bigr\rVert_1
\;+\;
\frac{\gamma}{M}
\sum_{j=1}^{M}
\bigl(1 - \mathrm{GIoU}(b_j,\hat b_{\sigma^*(j)})\bigr)
\end{equation}

\begin{equation}
L_{\mathrm{cls}}^{\mathrm{unmatch}}
= \frac{\alpha}{N - M}
\sum_{i\notin\sigma^*([1,M])}
L_{\mathrm{cls}}\bigl(\hat p_i,\,0\bigr)
\end{equation}
where, $\alpha$, $\beta$, and $\gamma$ denote the loss weights for the classification, $L_1$ loss, and GIoU \cite{rezatofighi2019generalized} loss, respectively. To speed up convergence, an auxiliary loss $\mathcal{L}_{\text {aux}}$ is introduced. Specifically, the core detection loss is applied to the first $L-1$ decoder layers, separately. The losses from these layers are then summed:
\begin{equation}
L_{\mathrm{aux}}
= \sum_{l=1}^{L-1} \mathcal{L}_{\mathrm{core}}^{(l)}
\end{equation}
where $\mathcal{L}_{\mathrm{core}}^{(l)}$ denotes the core detection loss computed on the $l$-th layer’s predictions.

$\mathcal{L}_{\mathrm{dn}}$ is a denoising loss, which aims to accelerate training convergence and stabilize the query learning process. For each training batch, we generate $K$ noisy copies of the ground‐truth boxes and class labels by applying small perturbations to the original annotations. Specifically, if each of the $M$ ground‐truth boxes is duplicated $G$ times, then
\begin{equation}
K = G \times M \
\end{equation}
where $G$ is 100 following the RT-DETR setting. Then, we denote these denoising queries by
\begin{equation}
\{\bigl(\,\hat b_k^{\mathrm{dn}},\,\hat p_k^{\mathrm{dn}}\bigr)\}_{k=1}^K
\end{equation}
which are fed into the decoder. Then apply the $\mathcal{L}_{\text {DETR}}$ to supervise the denoising outputs:
\begin{equation}
\mathcal{L}_{\mathrm{dn}}
= \mathcal{L}_{\text {DETR}}\bigl(\{\hat b_k^{\mathrm{dn}},\,\hat p_k^{\mathrm{dn}}\}_{k=1}^K,\;\{b_j,c_j\}_{j=1}^M\bigr)
\end{equation}
where the one‐to‐one matching between each noisy query $k$ and its ground‐truth target $j$ is determined by a static assignment method.

\subsubsection{Segmentation task} We individually design loss function for drivable area segmentation $\mathcal{L}_{\text {segda }}$ and lane line segmentation $\mathcal{L}_{\text {segll }}$.

For drivable area segmentation, due to its large and contiguous features, we use a Focal Loss \cite{lin2017focal} to handle foreground-background imbalance and a standard Binary Cross‐Entropy loss to enforce region consistency:
\begin{equation}
\mathcal{L}_{\text{segda}}
= \lambda_{\mathrm{fl}}\;L_{\mathrm{FL}}\bigl(\hat y_{\mathrm{da}},\,y_{\mathrm{da}}\bigr)
\;+\;\lambda_{\mathrm{bce}}\;L_{\mathrm{BCE}}\bigl(\hat y_{\mathrm{da}},\,y_{\mathrm{da}}\bigr)
\end{equation}
where \(\hat y_{\mathrm{da}}\) and \(y_{\mathrm{da}}\) denote the predicted logit map and ground‐truth mask for the drivable area, respectively. $\lambda_{\mathrm{fl}}$ and $\lambda_{\mathrm{bce}}$ are weights of the Focal Loss and Binary Cross‐Entropy loss, respectively.

For lane line segmentation, it comes with narrow and sparse. It is easy to miss detections. Therefore, we use Focal Loss \cite{lin2017focal} to handle foreground-background imbalance and Tversky Loss \cite{salehi2017tversky} to penalize false negatives strongly:
\begin{equation}
\mathcal{L}_{\text{segll}}
= \lambda_{\mathrm{fl}}\;L_{\mathrm{FL}}\bigl(\hat y_{\mathrm{ll}},\,y_{\mathrm{ll}}\bigr)
\;+\;\lambda_{\mathrm{tv}}\;L_{\mathrm{TV}}\bigl(\hat y_{\mathrm{ll}},\,y_{\mathrm{ll}}\bigr)
\end{equation}
where \(\hat y_{\mathrm{ll}}\) and \(y_{\mathrm{ll}}\) denote the predicted logit map and ground‐truth mask for lane lines, respectively. $\lambda_{\mathrm{fl}}$ and $\lambda_{\mathrm{tv}}$ are weights of the Focal Loss and Tversky loss, respectively.

\section{Experiments and Results}
In this section, we introduce our experiment details, including the dataset, evaluation metrics and setup. Then, we evaluate RMT-PPAD on the BDD100K dataset and compare it to open-source methods. We analyze performance based on quantitative and visualization results. We also present comprehensive ablation studies to evaluate the effectiveness of our proposed modules. Additionally, we present a visualization result on real-world scenarios. 

\subsection{Experiment details}
\subsubsection{Dataset} The BDD100K \cite{yu2020bdd100k} is a broadly used dataset for autonomous driving research. It includes 100K annotated images with support for multiple perception tasks. Additionally, BDD100K is notable for its diverse scene types and diverse weather conditions. This heterogeneity makes BDD100K particularly suitable to evaluate model performance applied in panoptic driving perception tasks. The dataset is split into three subsets: 70K images for training, 10K for validation, and 20K for testing. As the test set annotations are not available, we follow the setting in previous works \cite{wu2022yolop, wang2024you, vu2022hybridnets}. All evaluations are conducted on the validation set. Additionally, the object detection task focuses exclusively on the ``vehicle'' category, which includes cars, buses, trucks, and trains.

\subsubsection{Evaluation Metrics} 
For object detection, we adopt mAP50 and Recall as the metrics. mAP50 measures the average precision across all classes when the predicted bounding box overlaps the ground truth by at least 50\%. It reflects the model’s overall accuracy in both localization and classification. Recall indicates the model's ability to detect all relevant objects. It measures the ratio of true positives to the total number of ground-truth instances. For drivable area segmentation, we adopt mean intersection over union (mIoU) as the evaluation metric. It calculates the average overlap between predicted and ground-truth masks. For lane line segmentation, we have already discussed evaluation metrics in \ref{ll_lable_metrics}: IoU and ACC. 

Additionally, we include FPS metrics, which measure the model's speed and efficiency by calculating how many frames it can process per second. FPS also reflects the model's ability to handle tasks in real-time. A higher FPS indicates a faster model. FPS can be calculated as:
\begin{equation}
\text{FPS} = \frac{N}{t},
\end{equation}
where $N$ is the number of frames processed, and $t$ is the processing time.

\begin{table*}[ht]
\centering
\caption{Quantitative results comparison of RMT-PPAD and open-resource MTL models on BDD100K.}
\label{tab:comparison_bdd100k}
\renewcommand{\arraystretch}{1.2}
\begin{tabularx}{\textwidth}{c c c *{6}{>{\centering\arraybackslash}X}}
\toprule
\multirow{2}{*}{\textbf{Model}} & \multirow{2}{*}{\textbf{FPS}} & \multirow{2}{*}{\textbf{Params (M)}} &
\multicolumn{2}{c}{\textbf{Object Detection}} &
\multicolumn{1}{c}{\textbf{Drivable Area}} &
\multicolumn{2}{c}{\textbf{Lane Line}} \\
\cmidrule(lr){4-5} \cmidrule(lr){6-6} \cmidrule(lr){7-8}
& & & \textbf{Recall (\%)} & \textbf{mAP50 (\%)} & \textbf{mIoU (\%)} & \textbf{IoU (\%)} & \textbf{ACC (\%)} \\
\midrule
YOLOP       & \textbf{64.5}  & 7.9  & 88.5 & 76.4 & 89.0  & 44.0 & 79.8 \\
HybridNet   & 17.2  & 12.8 & 93.5 & 77.2 & 91.0  & 52.0 & 82.7 \\
YOLOPX      & 27.5  & 32.9 & 93.7 & 83.3 & 90.9  & 52.1 & 79.1 \\
A-YOLOM(n)  & 52.9 & \textbf{4.4}  & 85.3 & 78.0 & 90.5  & 45.6 & 77.2 \\
A-YOLOM(s)  & 52.7  & 13.6 & 86.9 & 81.1 & 91.0  & 49.7 & 80.7 \\
RMT-PPAD        & 32.6  & 34.3 & \textbf{95.4} & \textbf{84.9} & \textbf{92.6}  & \textbf{56.8} & \textbf{84.7} \\
\bottomrule
\end{tabularx}
\end{table*}

\subsubsection{Experimental Setup and Implementation}
We compare RMT-PPAD with the open-source MTL methods in panoptic driving perception tasks. Specifically, we compare our model with YOLOPX \cite{zhan2024yolopx}, A-YOLOM(n), A-YOLOM(s) \cite{wang2024you}, HybridNet \cite{vu2022hybridnets}, and YOLOP \cite{wu2022yolop}. Since we correct the label of lane line segmentation and for fair comparison, we can only reproduce the open-source methods to compare. 

We train our model using the SGD optimizer with a learning rate (lr) of 0.01, momentum of 0.9, and a weight decay of 0.0005. We begin training with a warm-up for 3 epochs. During this stage, the momentum of the SGD optimizer is set to 0.8, and the learning rate for biases is initialized at 0.1. A cosine learning rate schedule is used to control the learning rate during training, enabling gradual and smooth decay. Additionally, we resize the input images from 1280 × 720 to 640 × 640. For loss function coefficients, we set $\alpha=1$, $\beta=5$, $\gamma=2$, $\lambda_{\mathrm{fl}}=24$, $\lambda_{\mathrm{bce}}=8$, and $\lambda_{\mathrm{tv}}=8$. During inference, we apply segmentation mask thresholds of 0.45 for drivable area and 0.9 for lane line. 

We train the model with a batch size of 45 for 250 epochs on three RTX 4090 GPUs around 70 hours. Due to the large scale of the BDD100K dataset, training a model requires a long time. Therefore, we perform ablation experiments on a toy dataset. This toy dataset consists of 10K samples randomly selected from the original BDD100K training set and 2K samples from the validation set. All training configurations are kept the same, except the number of training epochs is reduced to 200. Training on this toy dataset takes approximately 8 hours. 

All FPS-related evaluation experiments are performed on a single RTX 4090 GPU without inference accelerators, such as TensorRT or ONNX Runtime. And setting the batch size is 1.

\begin{figure*}[ht]
  \centering
  \begin{tabular}{l @{\quad} cccc}
    \rotatebox{90}{\textbf{YOLOP}} &
      \includegraphics[width=0.22\linewidth]{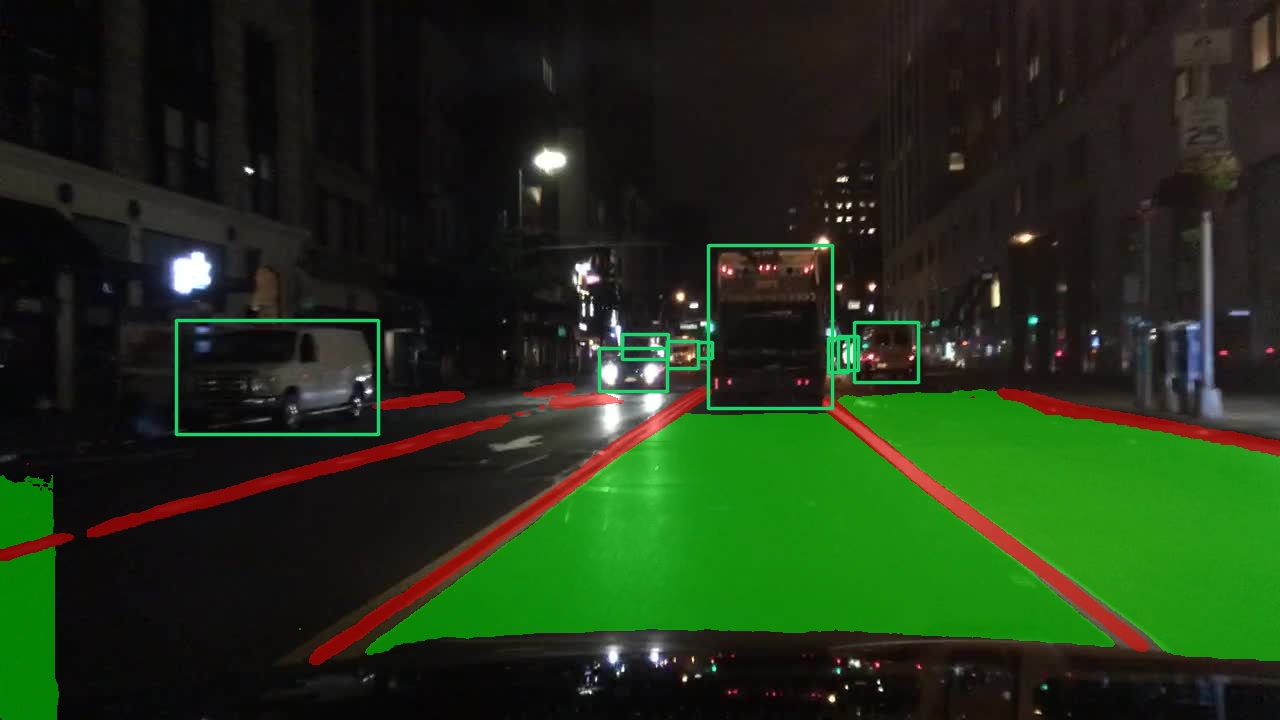} &
      \includegraphics[width=0.22\linewidth]{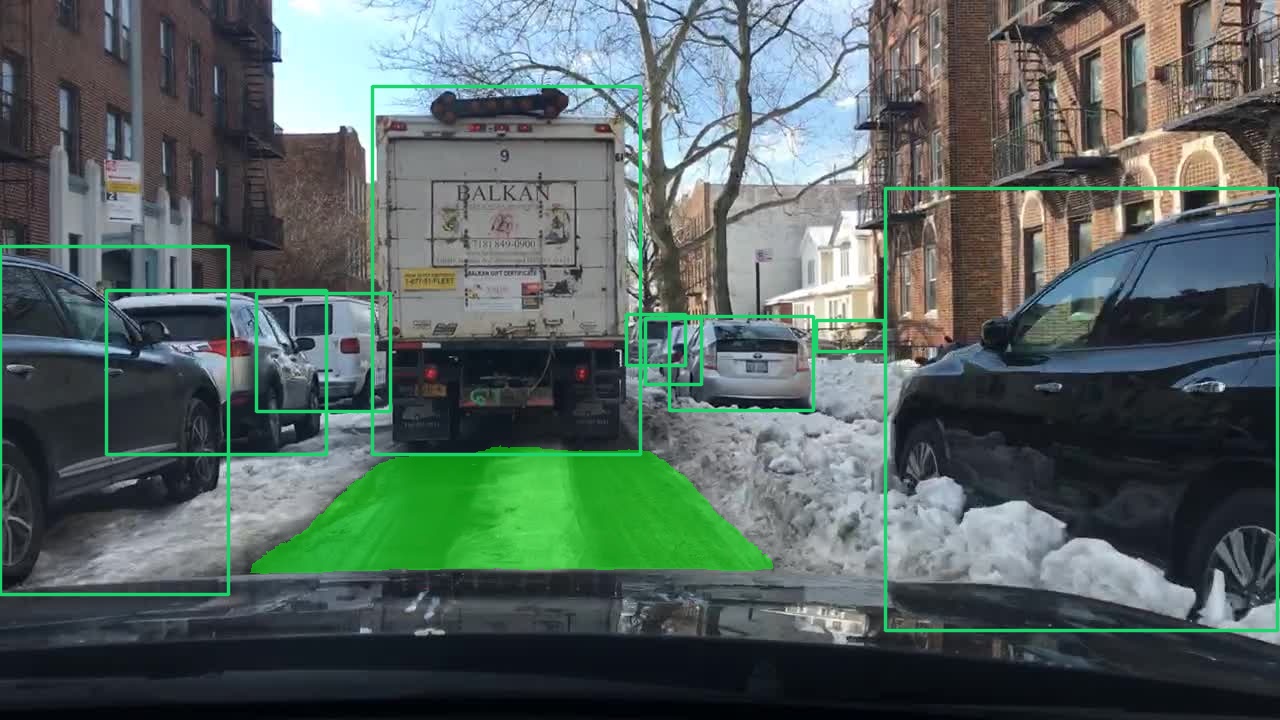} &
      \includegraphics[width=0.22\linewidth]{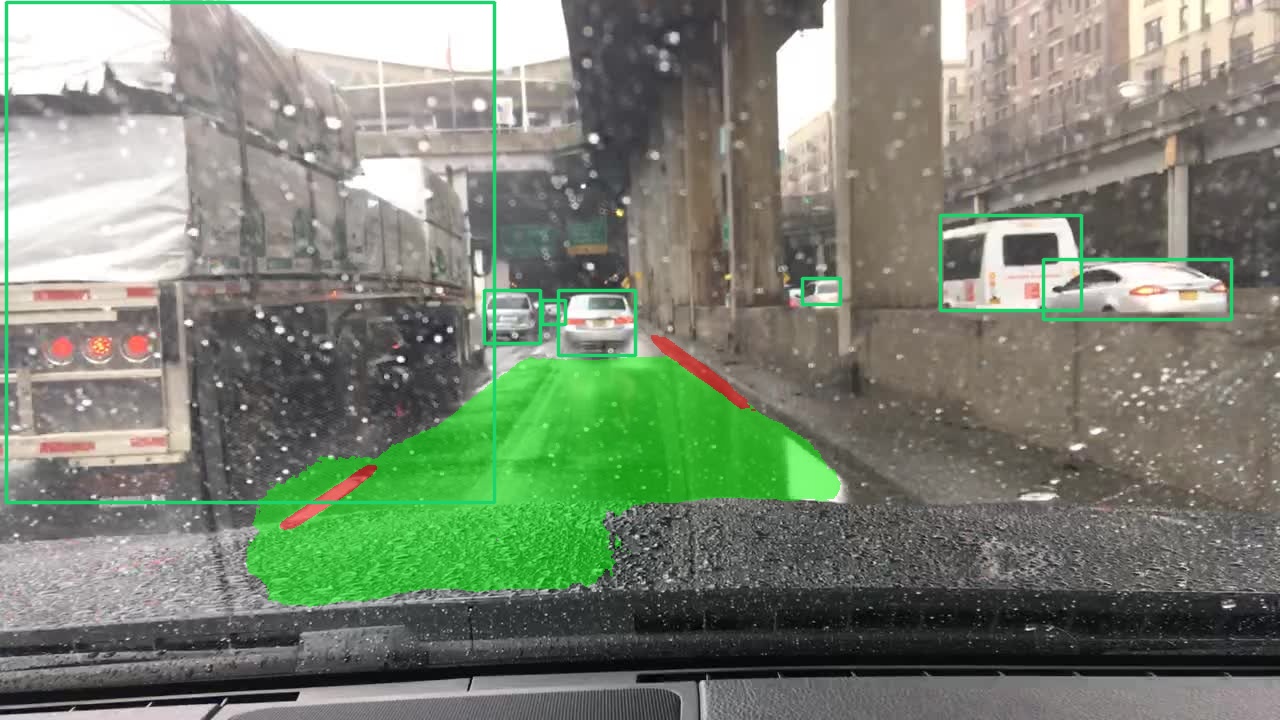} &
      \includegraphics[width=0.22\linewidth]{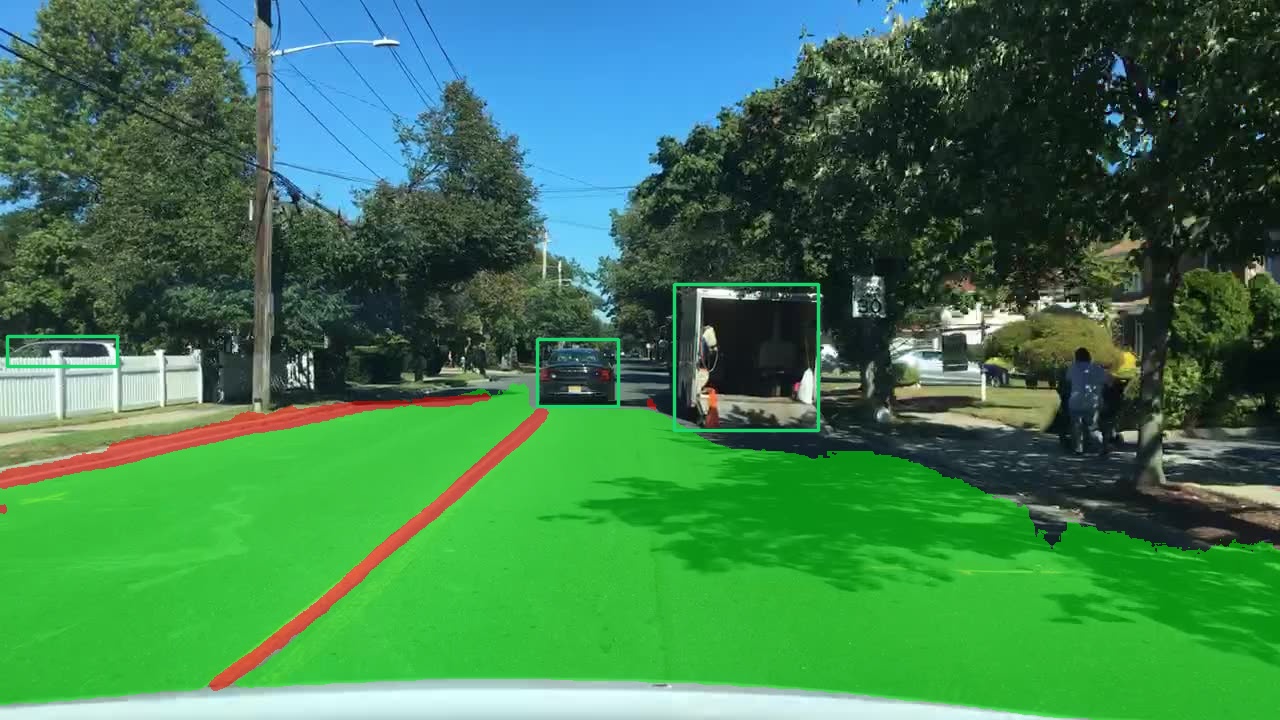} \\[6pt]

    \rotatebox{90}{\textbf{HybridNet}} &
      \includegraphics[width=0.22\linewidth]{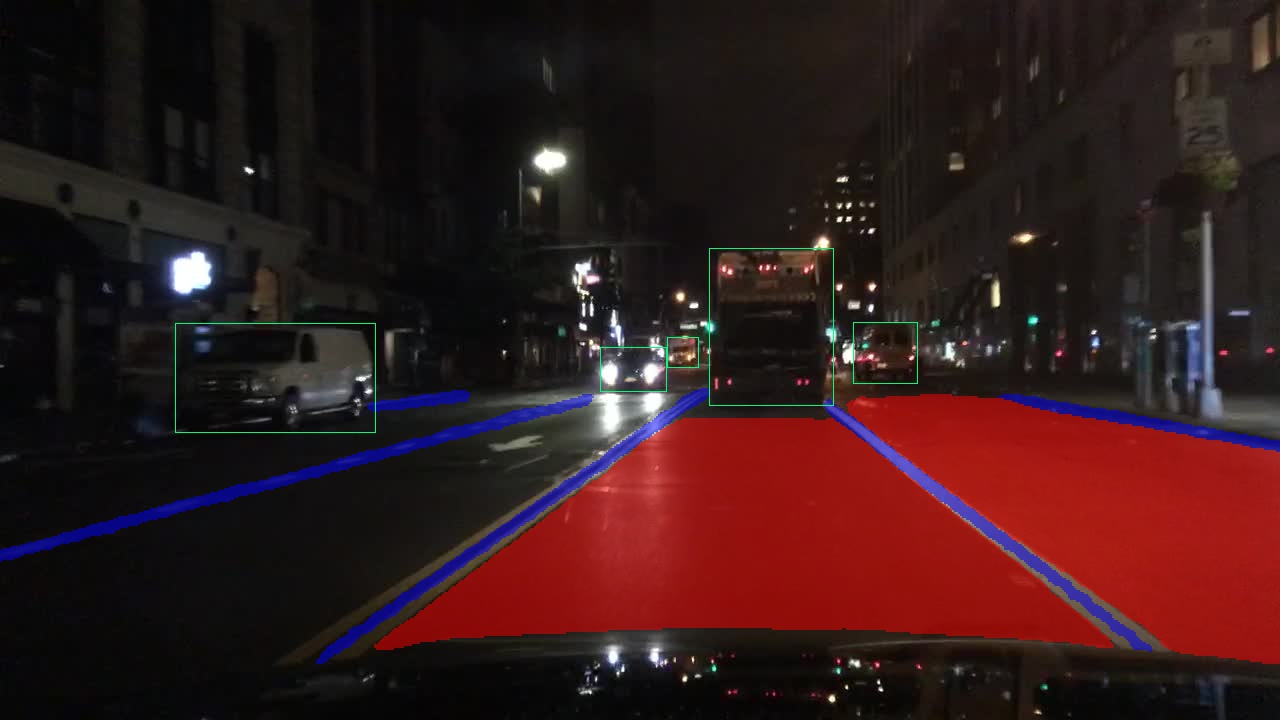} &
      \includegraphics[width=0.22\linewidth]{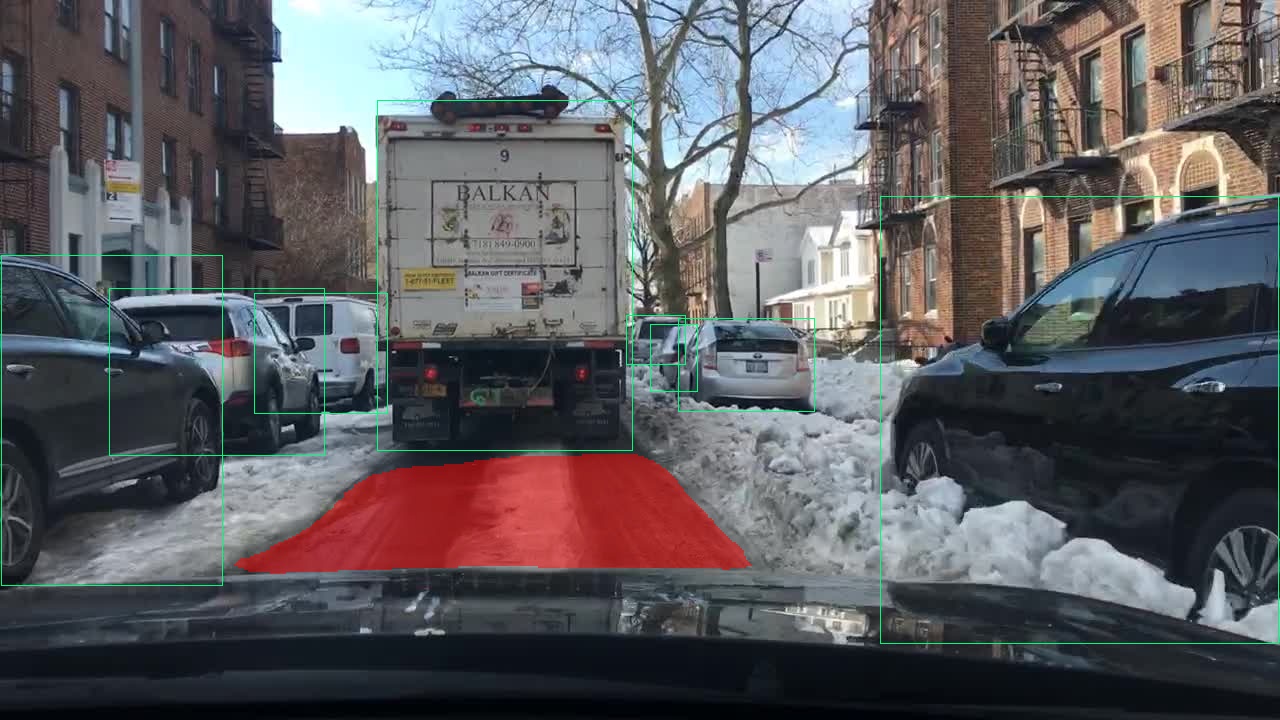} &
      \includegraphics[width=0.22\linewidth]{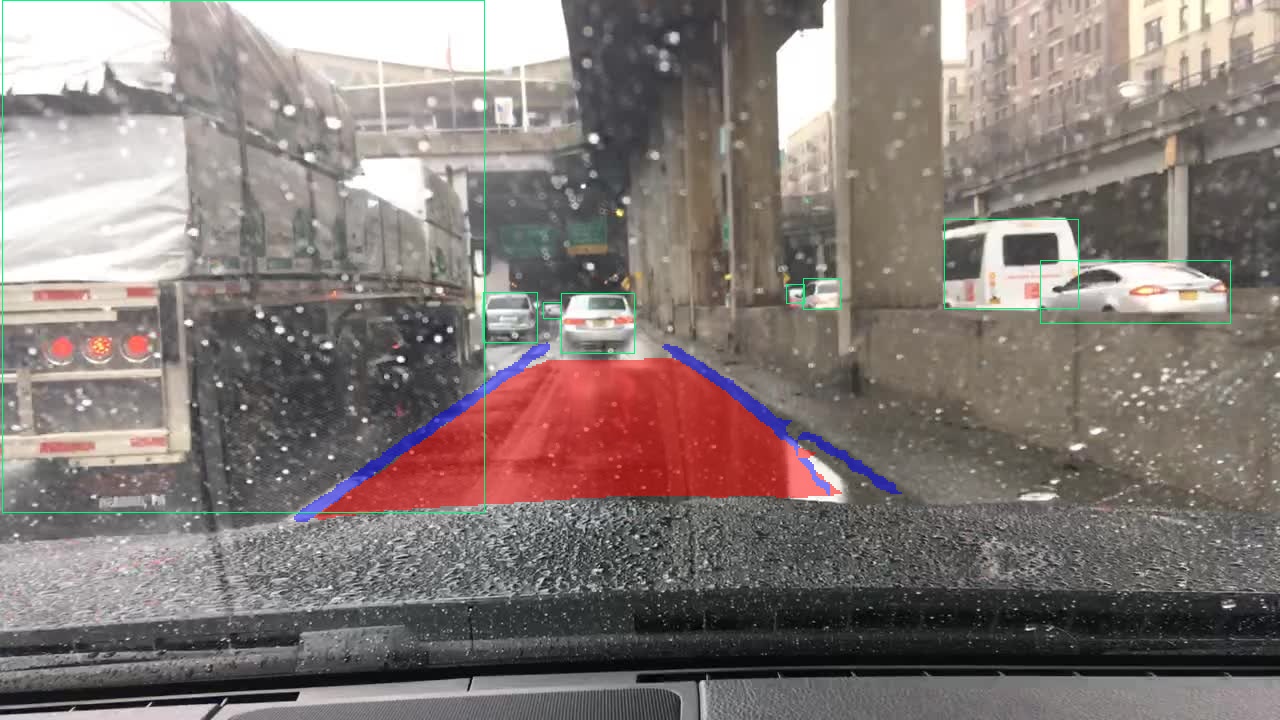} &
      \includegraphics[width=0.22\linewidth]{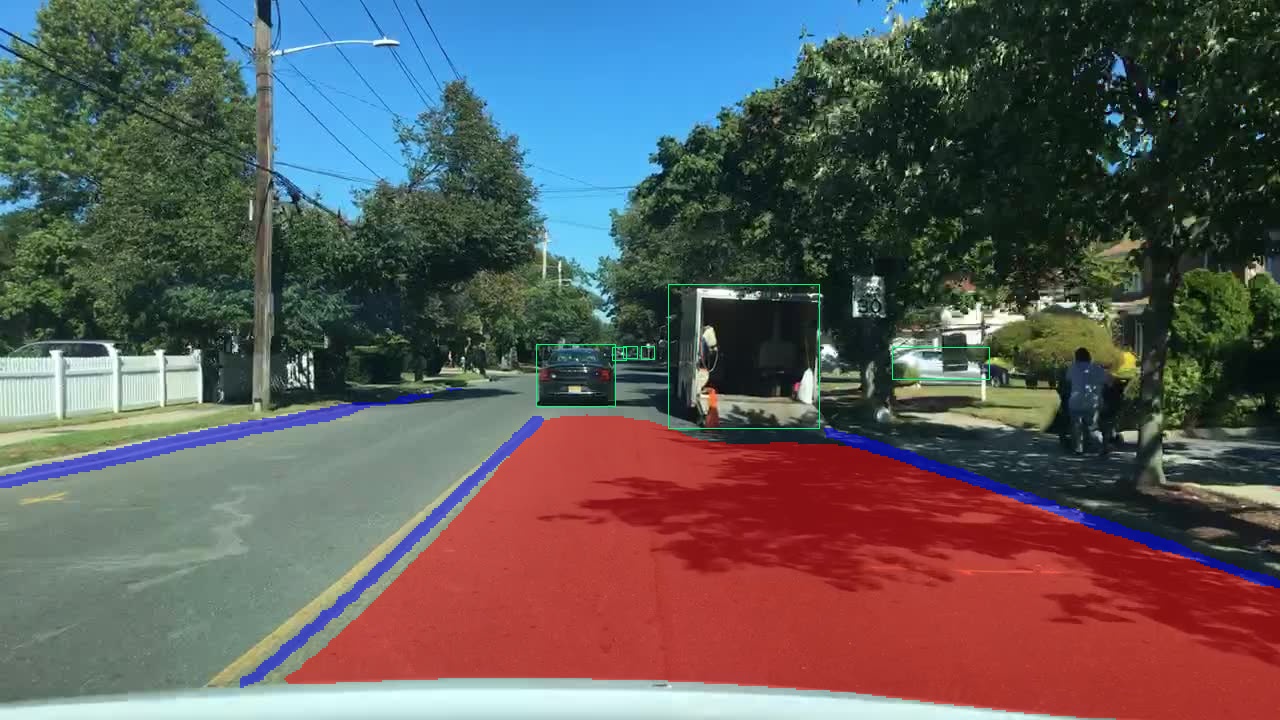} \\[6pt]

    \rotatebox{90}{\textbf{YOLOPX}} &
      \includegraphics[width=0.22\linewidth]{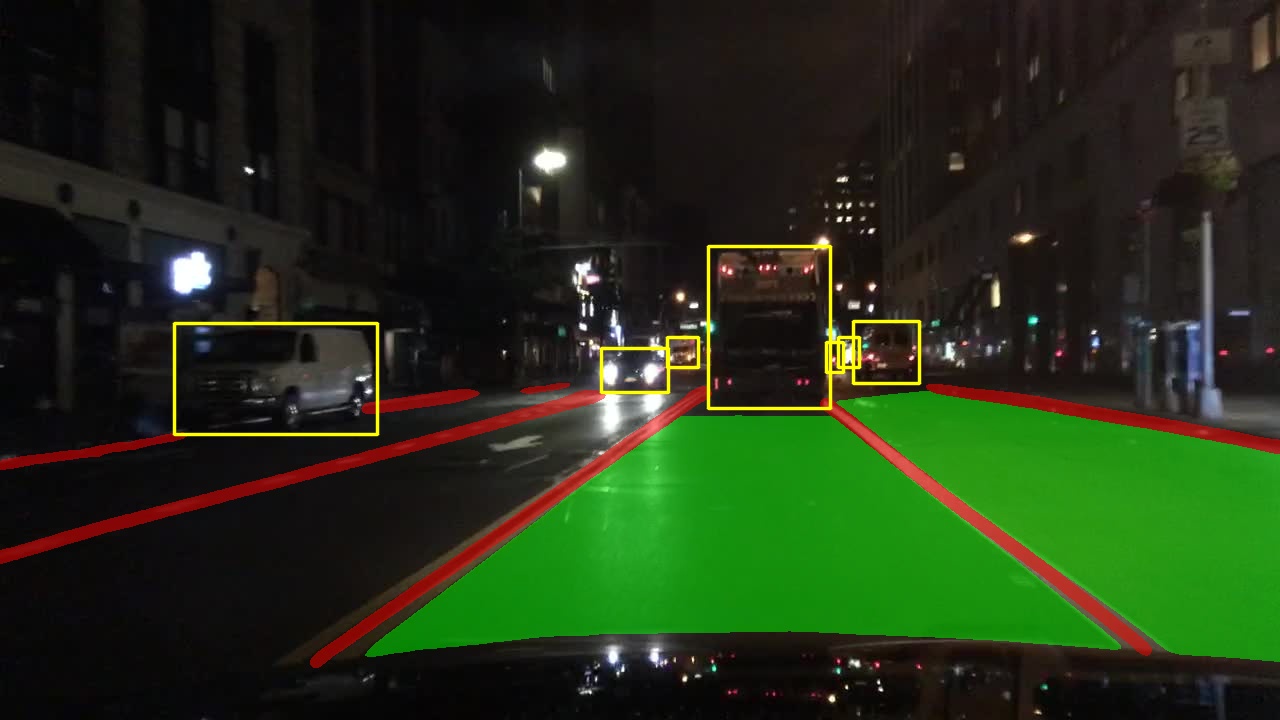} &
      \includegraphics[width=0.22\linewidth]{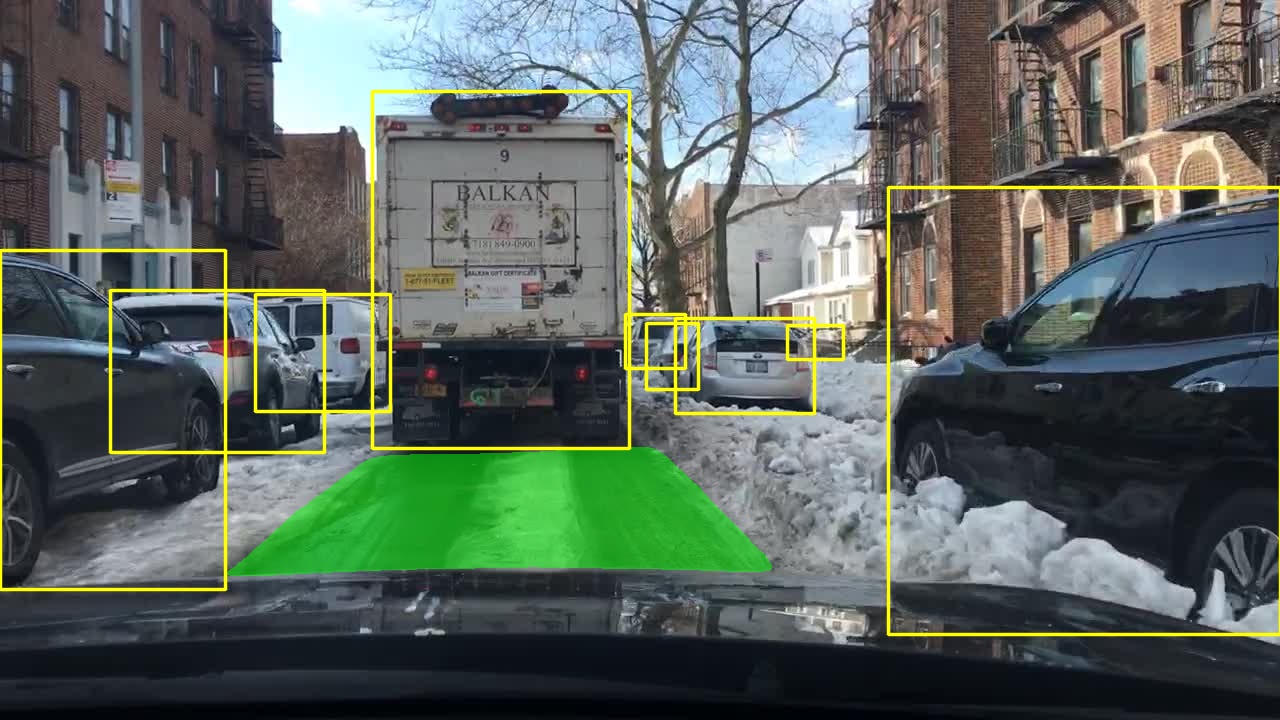} &
      \includegraphics[width=0.22\linewidth]{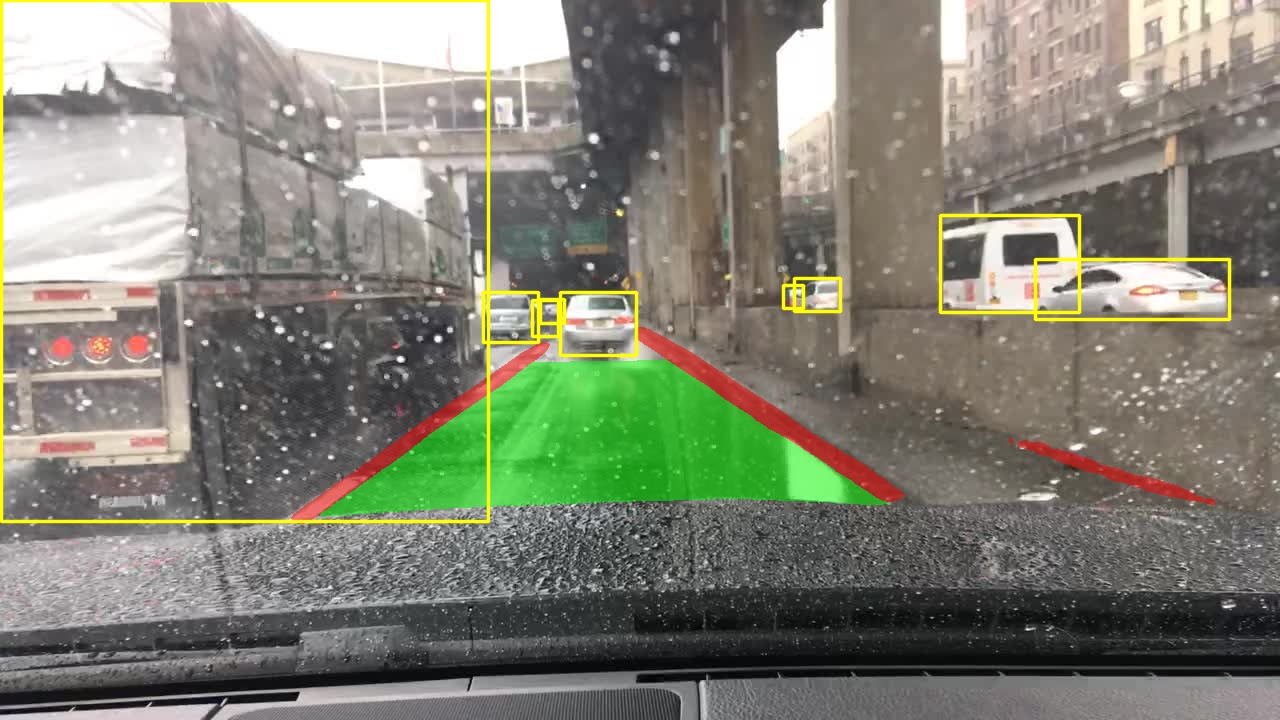} &
      \includegraphics[width=0.22\linewidth]{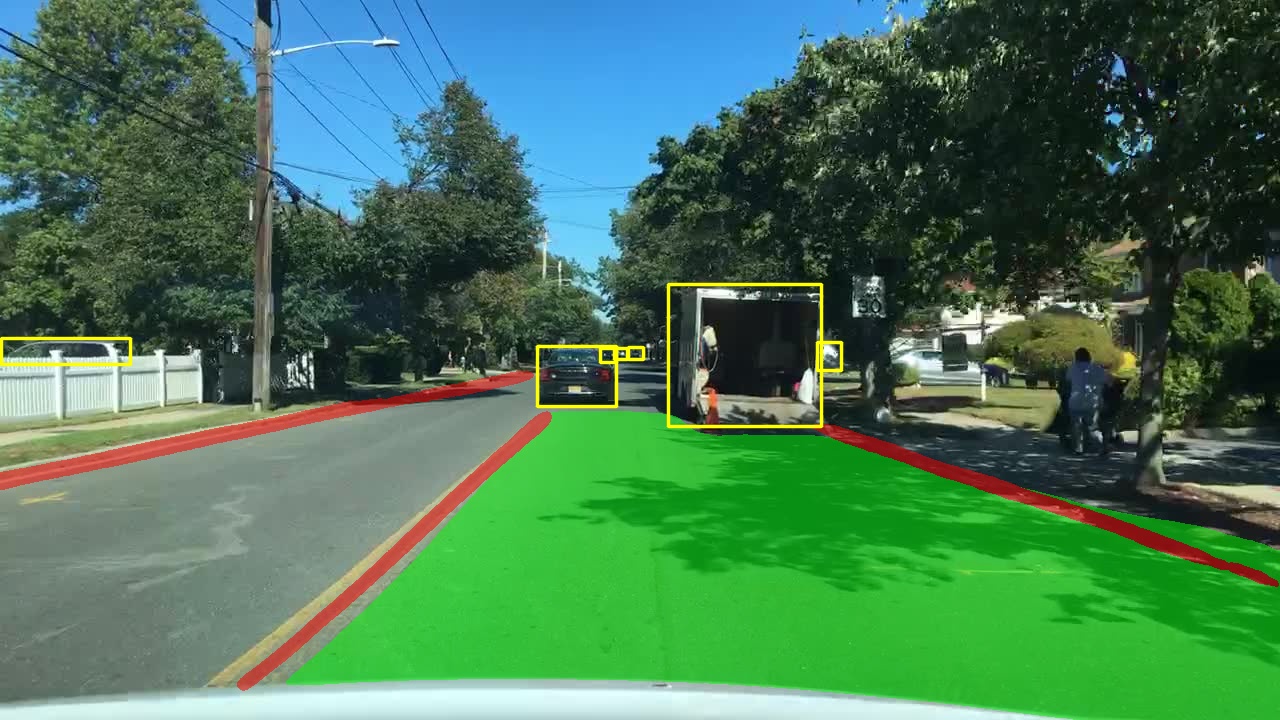} \\[6pt]

    \rotatebox{90}{\textbf{A-YOLOM(n)}} &
      \includegraphics[width=0.22\linewidth]{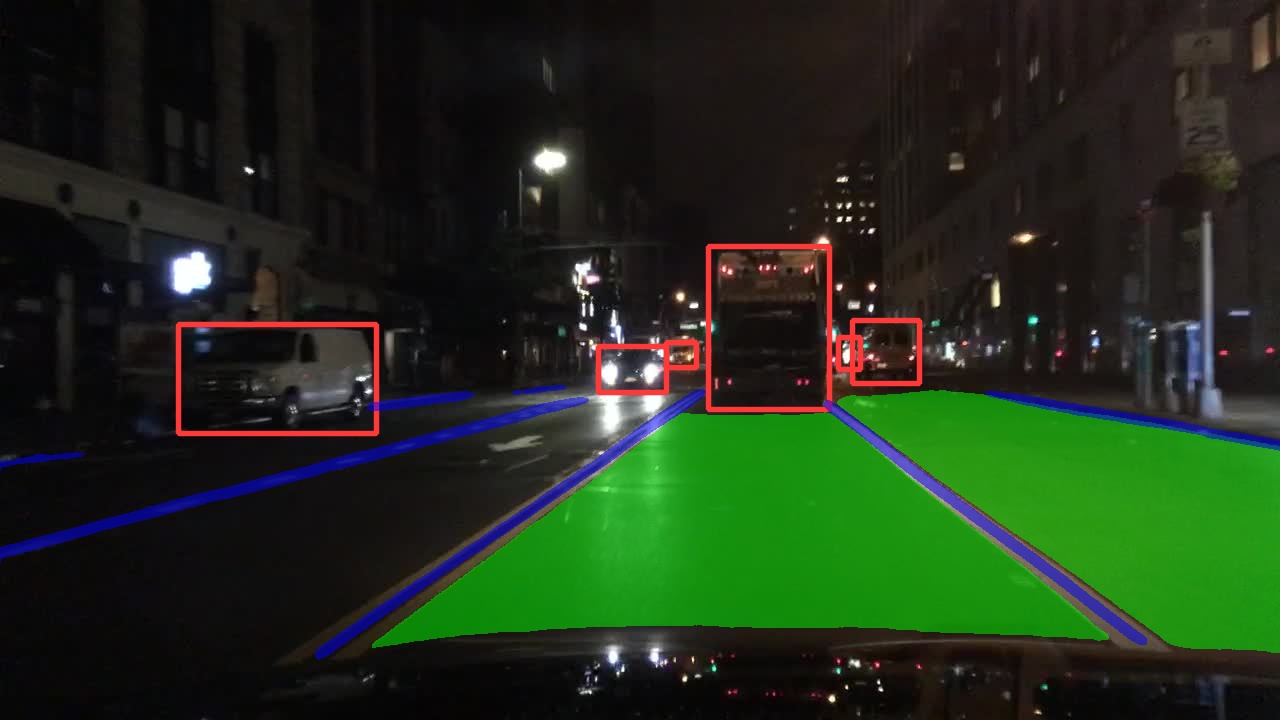} &
      \includegraphics[width=0.22\linewidth]{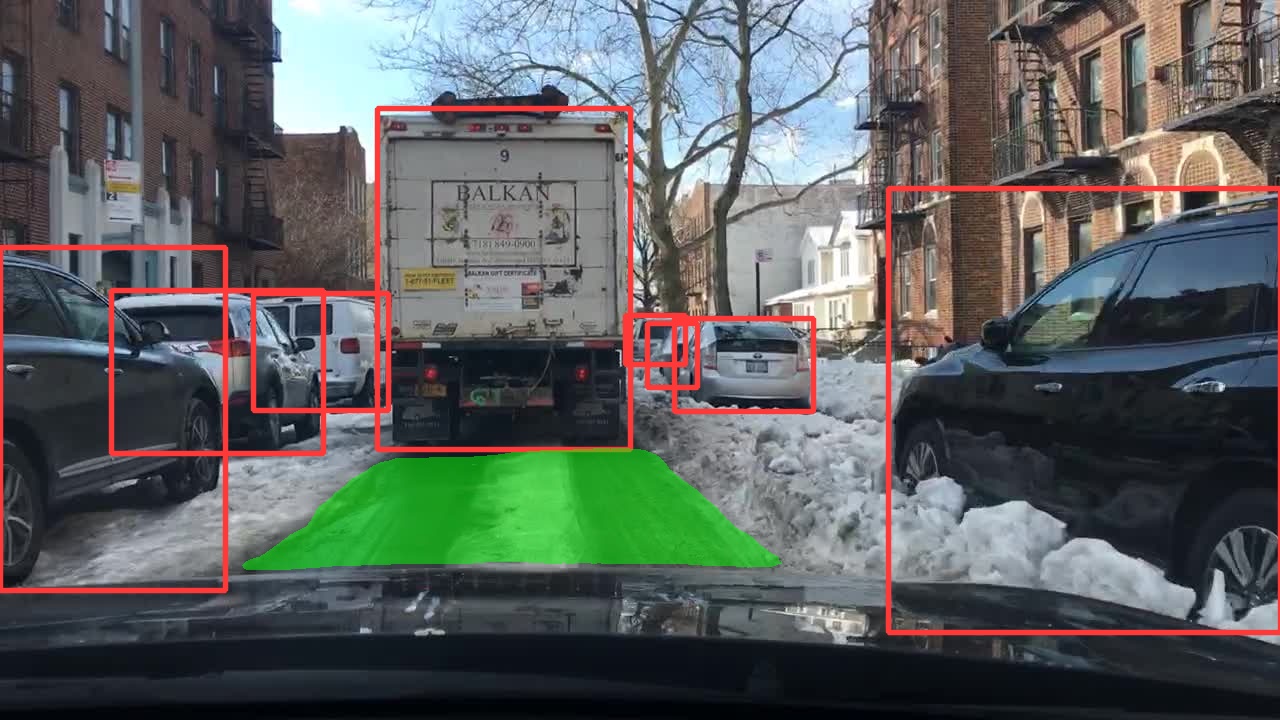} &
      \includegraphics[width=0.22\linewidth]{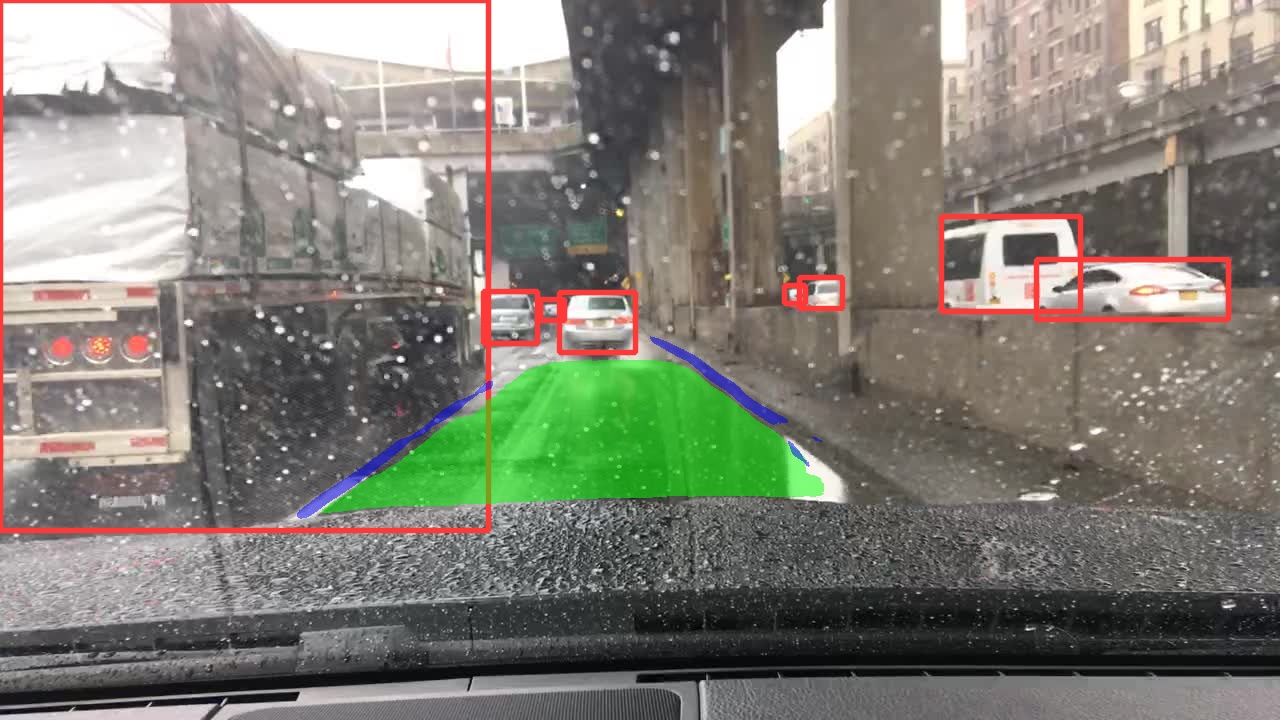} &
      \includegraphics[width=0.22\linewidth]{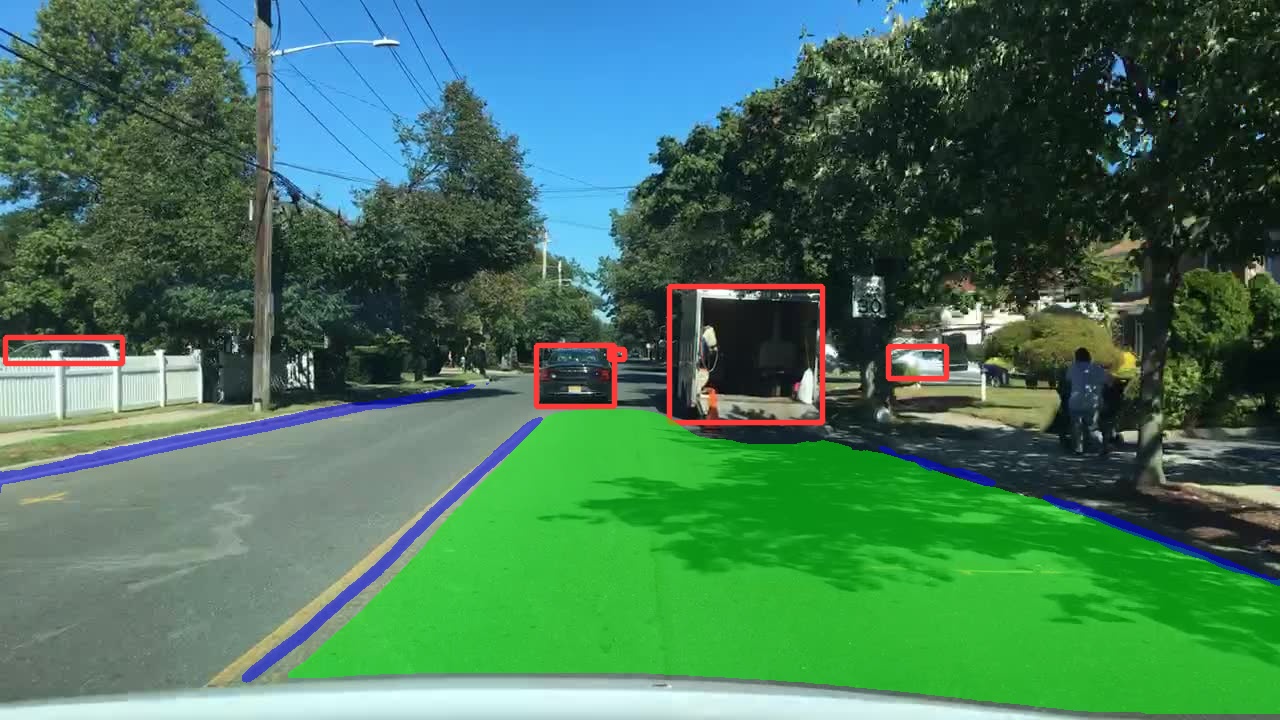} \\[6pt]

    \rotatebox{90}{\textbf{A-YOLOM(s)}} &
      \includegraphics[width=0.22\linewidth]{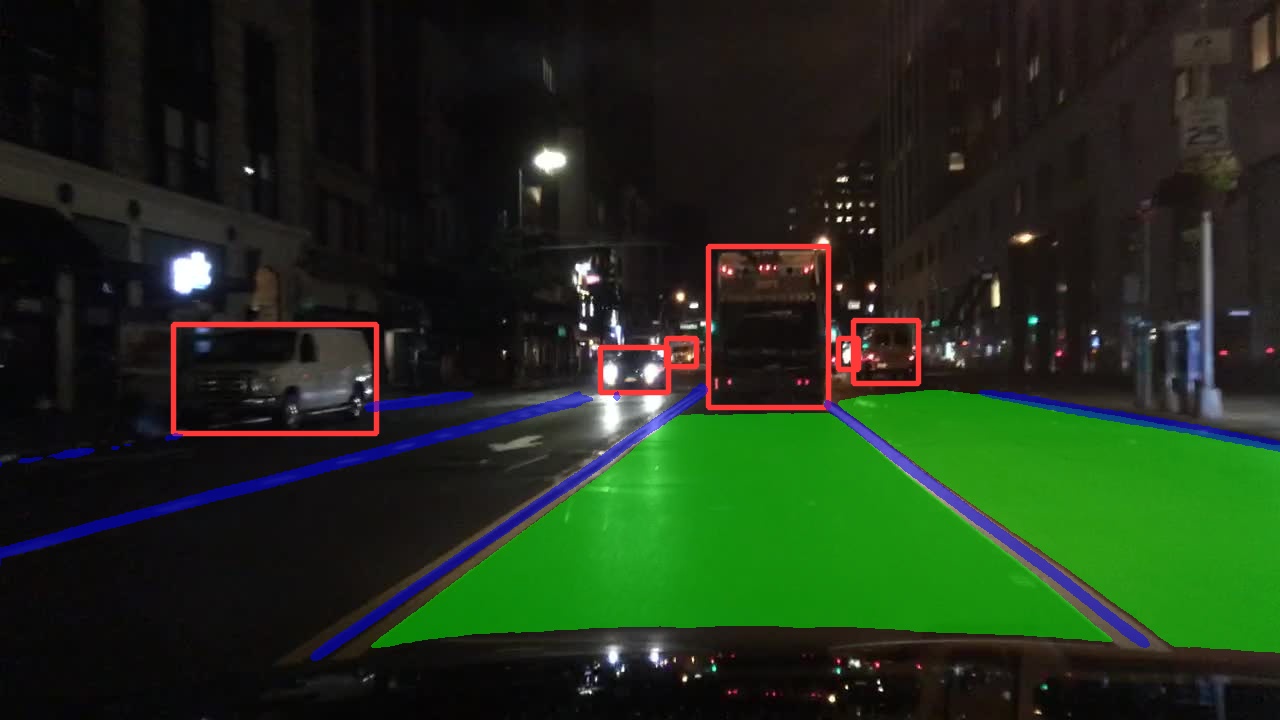} &
      \includegraphics[width=0.22\linewidth]{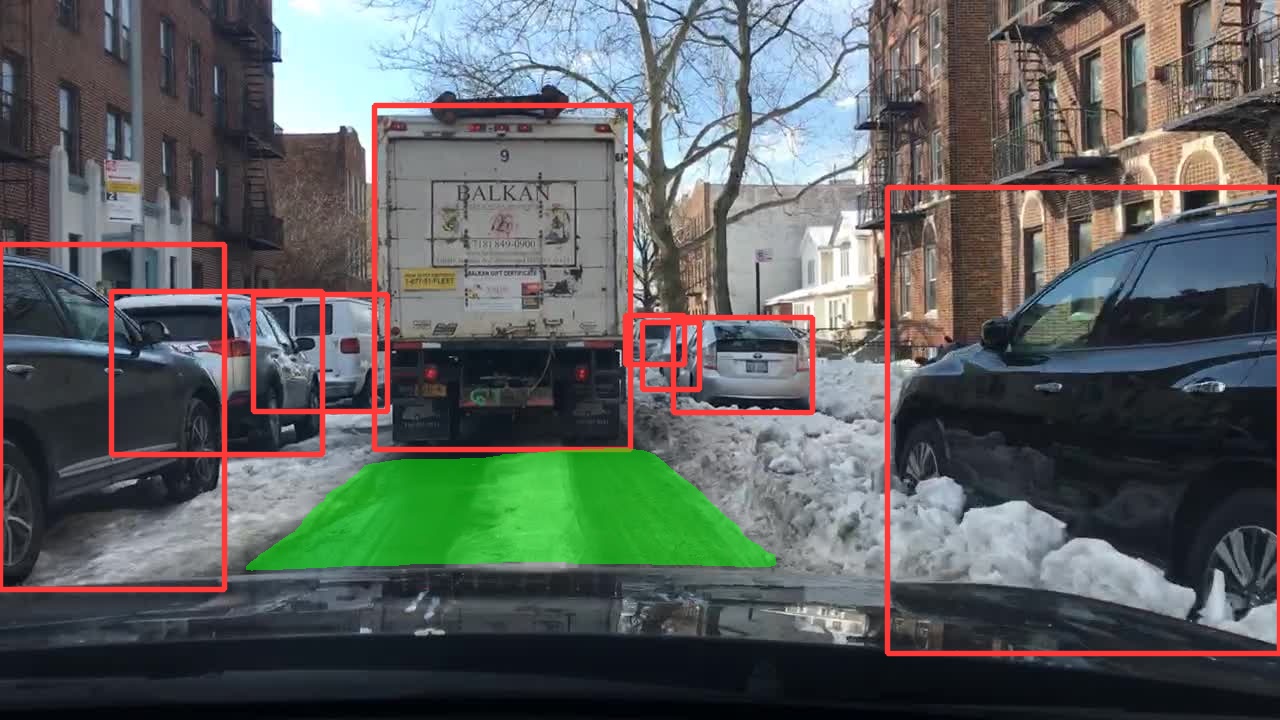} &
      \includegraphics[width=0.22\linewidth]{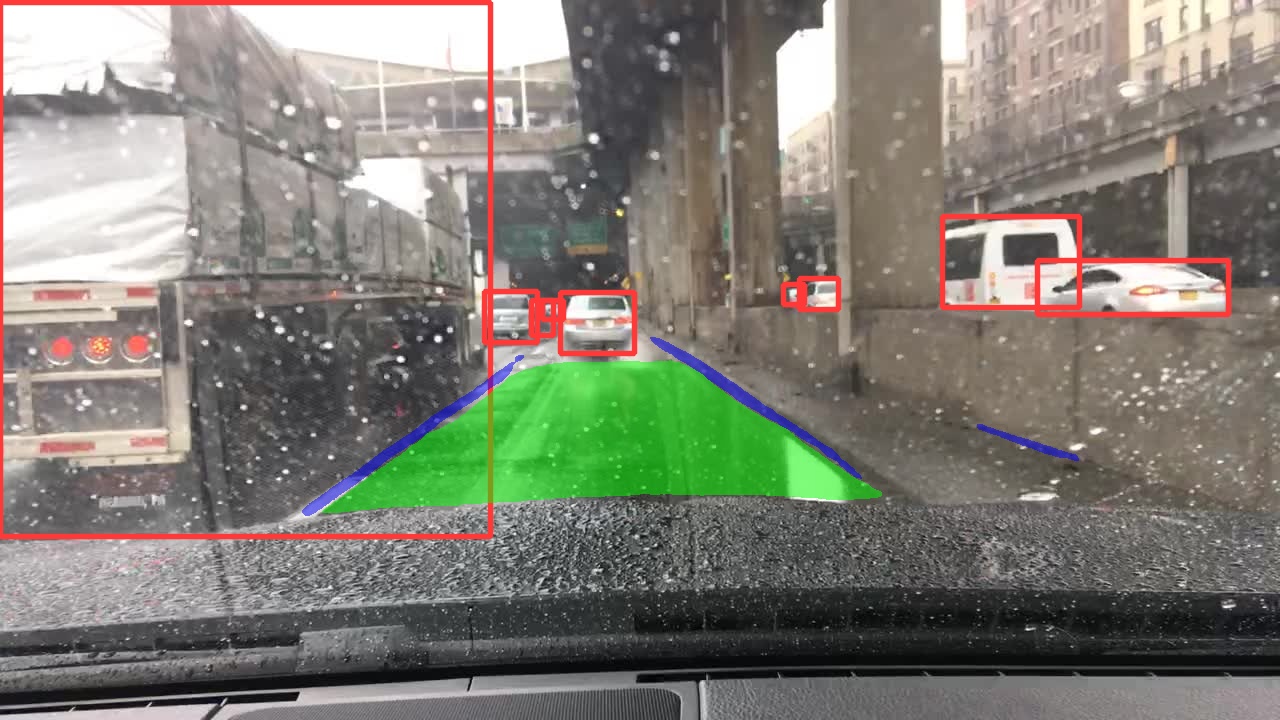} &
      \includegraphics[width=0.22\linewidth]{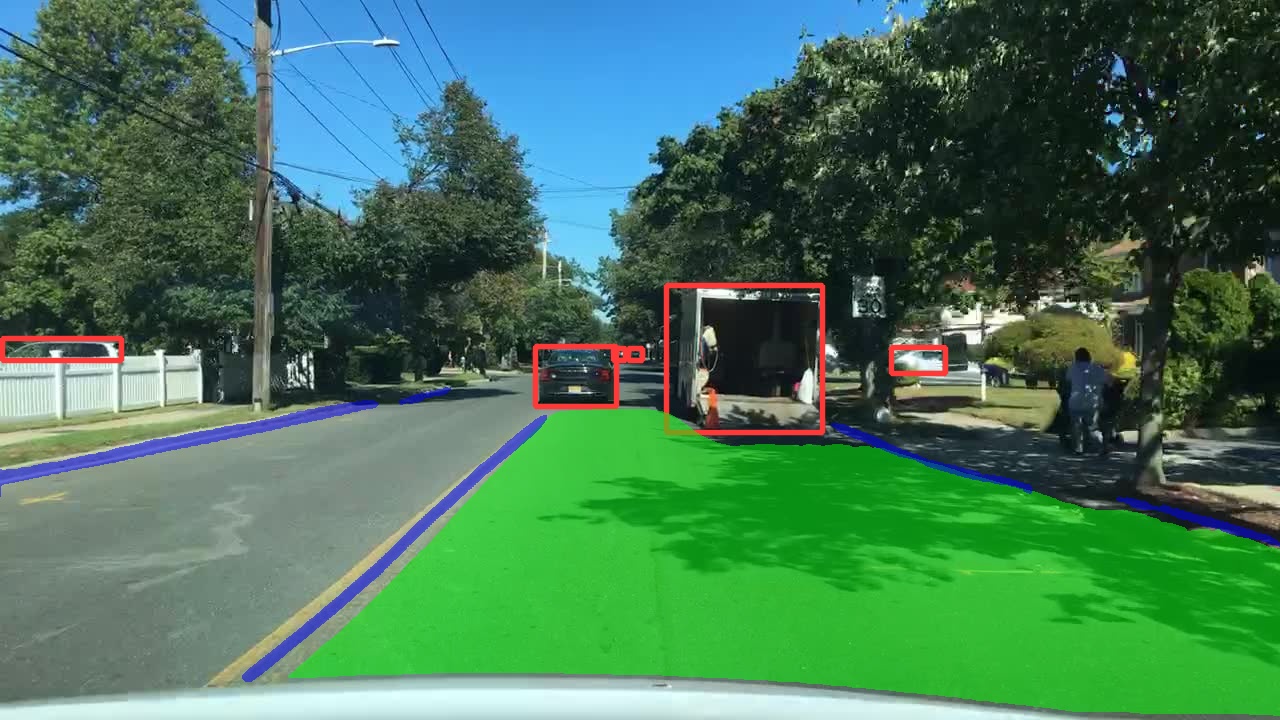} \\[6pt]

    \rotatebox{90}{\textbf{RMT-PPAD}} &
      \includegraphics[width=0.22\linewidth]{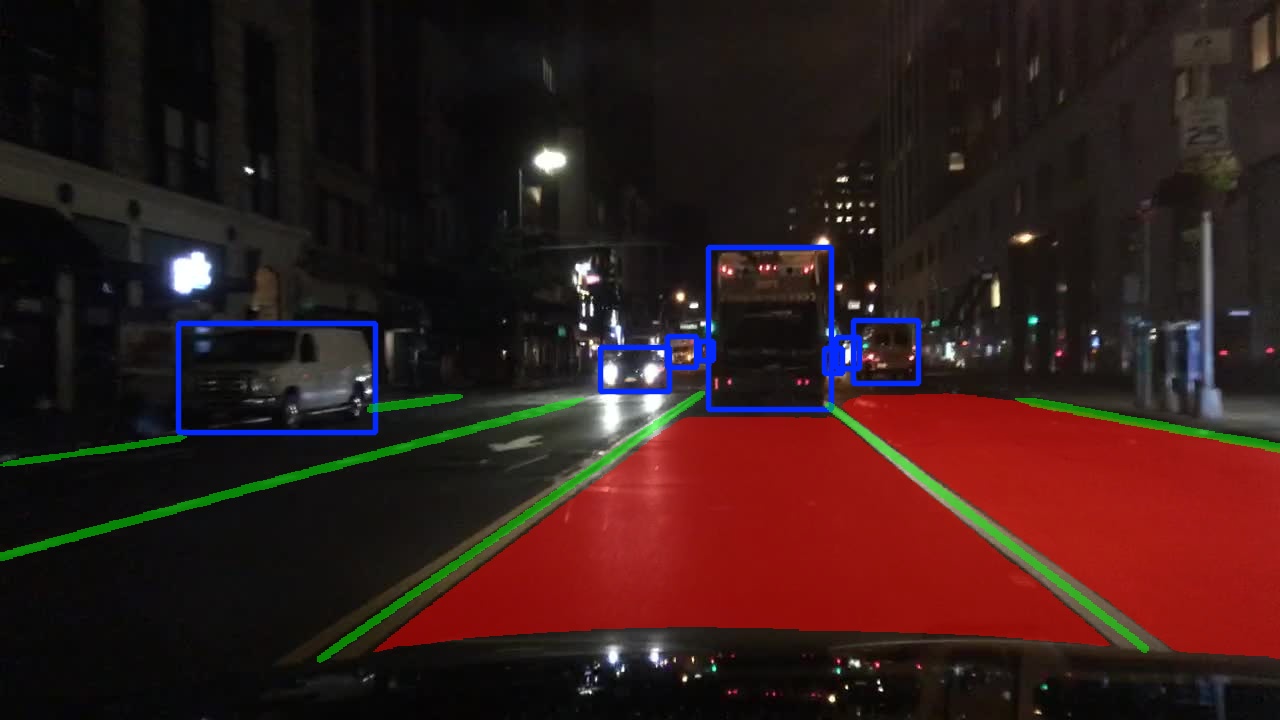} &
      \includegraphics[width=0.22\linewidth]{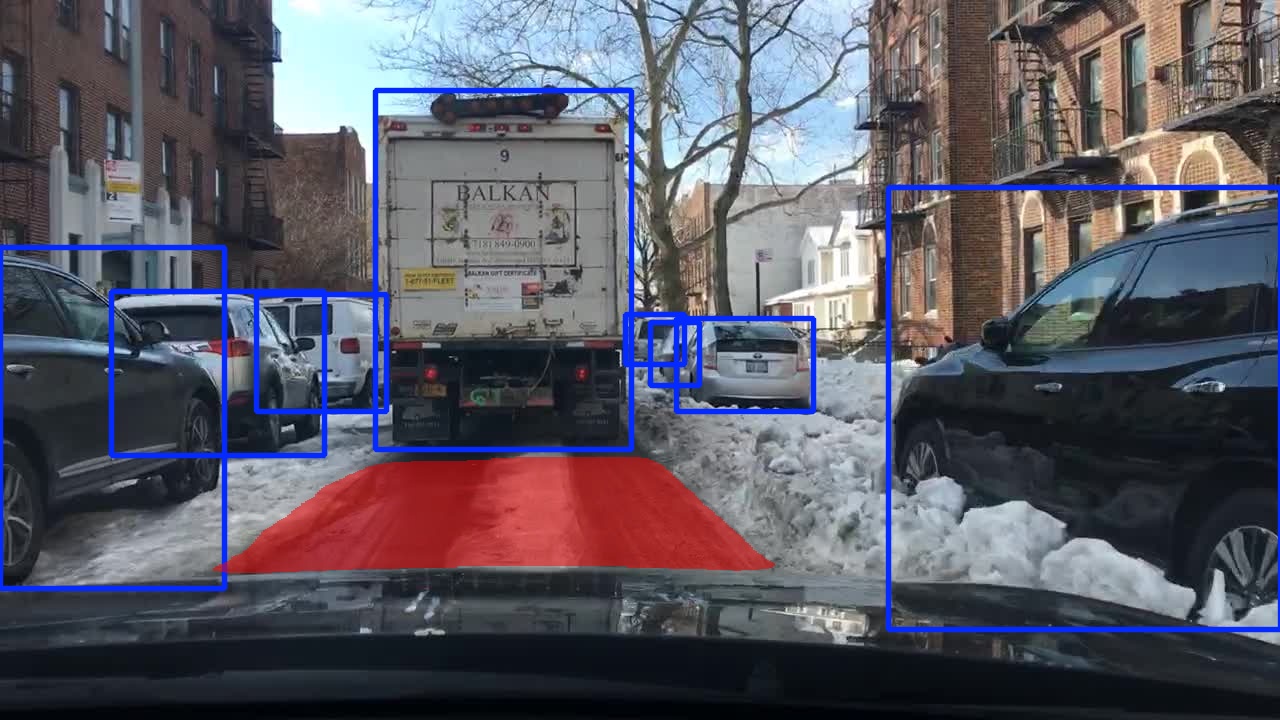} &
      \includegraphics[width=0.22\linewidth]{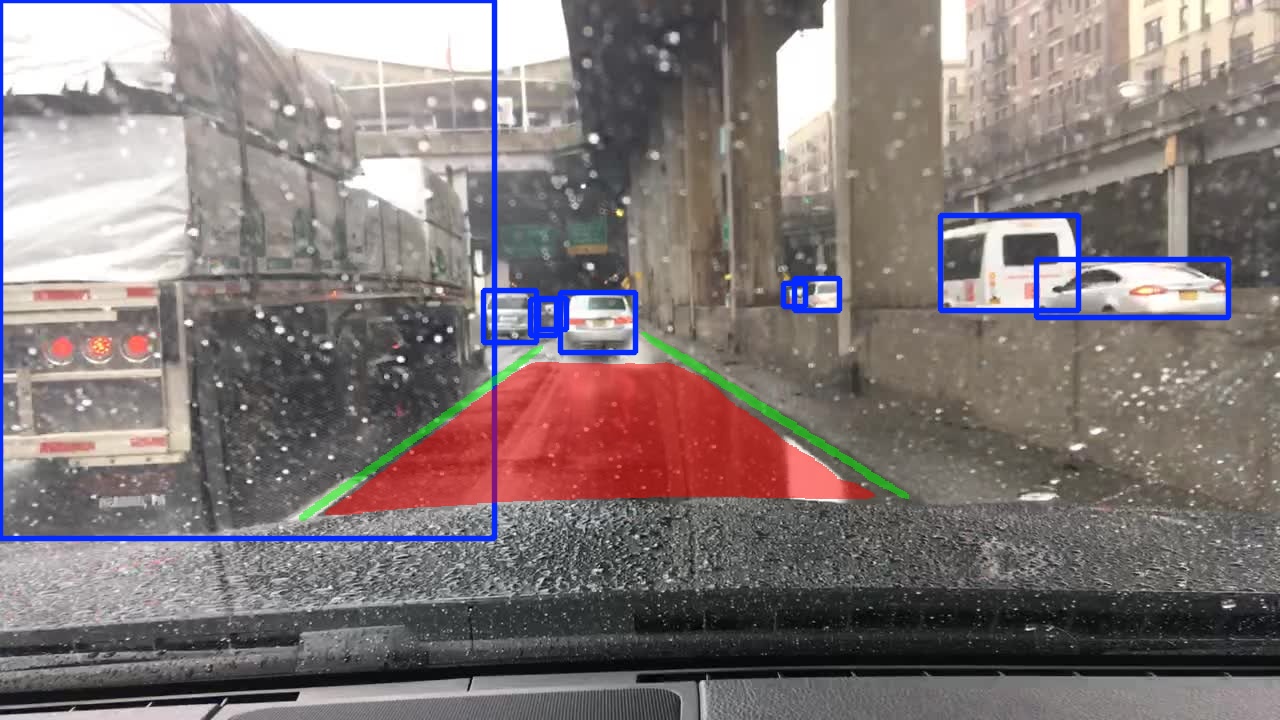} &
      \includegraphics[width=0.22\linewidth]{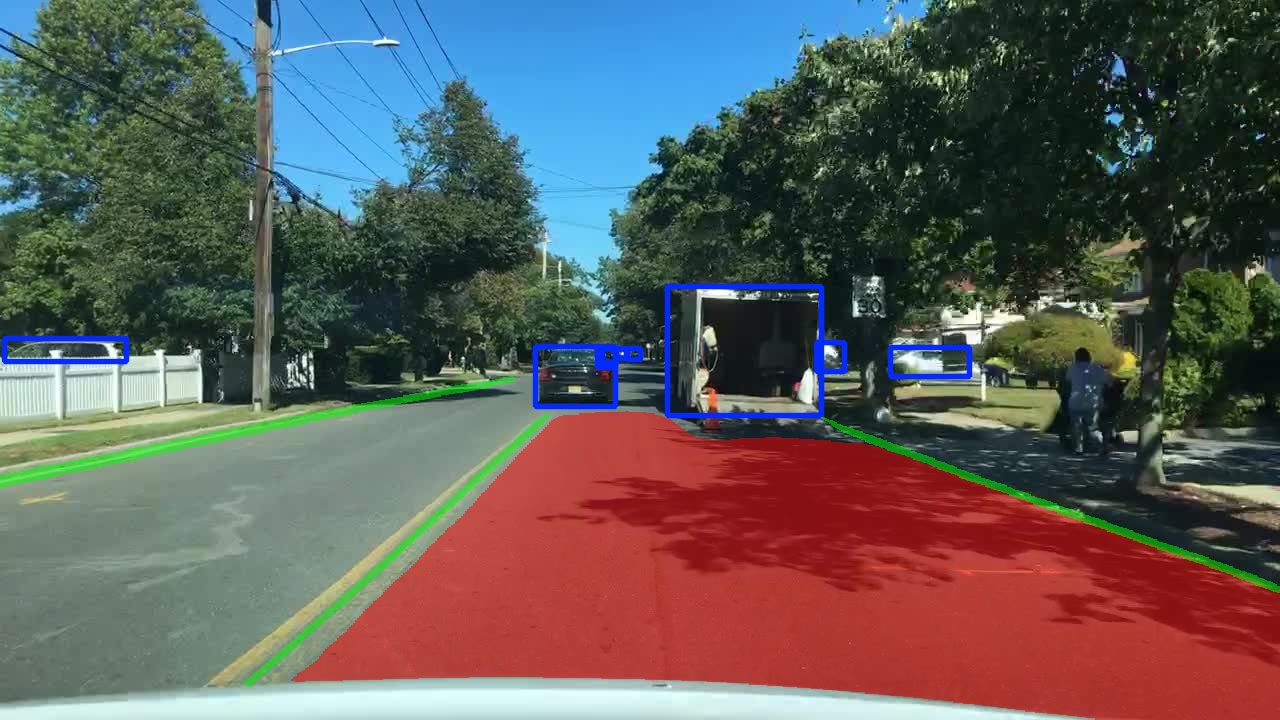} \\
  \end{tabular}
  \caption{Visualization results comparison on different scenarios. From left to right are night, snow, rain, and daytime. }
  \label{fig:vision_result}
\end{figure*}

\subsection{Experimental results}

\subsubsection{Quantitative results} 
We evaluate RMT-PPAD and reproduce YOLOP\footnote{https://github.com/hustvl/YOLOP}, YOLOPX\footnote{https://github.com/jiaoZ7688/YOLOPX}, HybridNet\footnote{https://github.com/datvuthanh/HybridNets}, A-YOLOM(n), and A-YOLOM(s)\footnote{https://github.com/JiayuanWang-JW/YOLOv8-multi-task}. The results are shown in Table \ref{tab:comparison_bdd100k}, RMT-PPAD achieves competitive efficiency and SOTA performance for three tasks on the BDD100K dataset. Specifically, RMT-PPAD achieves the best mAP50 of 84.9\% and Recall of 95.4\% in object detection. This outperforms all other open-source models. For drivable area segmentation, RMT-PPAD achieves the best mIoU of 92.6\%, indicating superior performance in road understanding. In lane line segmentation, RMT-PPAD achieves the best IoU of 56.8\% and ACC of 84.7\%. Like we discussed in Section \ref{ll_lable_metrics}, ACC can not penalize over-segmentation. Therefore, IoU will more accurately reflect the real performance of the model because it not only considers the $TP$ and $FN$, but also $FP$. 

Although our model includes a relatively larger number of parameters, it still achieves real-time inference and 32.6 FPS. Due to limitations in current sensors, most cameras operate at a sampling rate of 30 FPS. Therefore, any model that achieves inference speed above 30 FPS can be considered real-time. 

\subsubsection{Visualization results} 
Fig. \ref{fig:vision_result} shows the visualization results. We present a comparison of different methods under diverse driving scenarios, including night, snow, rain, and daytime conditions. Compared to other methods, RMT-PPAD consistently delivers more accurate and robust segmentation and detection results across all scenarios. 

In the night scenario, RMT-PPAD effectively segments the drivable area and lane lines. While other methods either produce parts of the wrong drivable area or produce wrong, blurry and fragmented lane lines, especially for YOLOP, HybridNet and A-YOLOM. In the snow scenario, RMT-PPAD still shows excellent detection performance. While YOLOP and YOLOPX fail to distinguish vehicles from the snow-covered background, leading to incorrect identification of part of the building as a vehicle. In the rain scenario, raindrops on the camera lens add noise to the captured images. This destroys fine-grained features and makes the segmentation task more challenging. RMT-PPAD handles rainy conditions effectively. It produces more accurate and smoother lane line segmentation. Other methods often show fragmented or imprecise lane markings in rainy scenes. Additionally, they also struggle to maintain accurate drivable area segmentation, especially for YOLOP. Finally, in the daytime scenario, most methods perform well. However, RMT-PPAD shows better robustness in challenging cases. Specifically, our method could accurately detect partially occluded vehicles. Other methods often miss these vehicles. Additionally, our method is also unaffected by tree shadows. Other methods produce fragmented lane lines or incorrect drivable areas under shadows.

\subsubsection{learnable weight results in the segmentation decoder} 
After training, we check the learned weights $\alpha$ for multi-scale feature maps [S3, S4, and F5] in the segmentation decoder. For the drivable area segmentation, the weights are [0.355, 0.156, 0.489]. For the lane line segmentation, the weights are [0.405, 0.442, 0.153]. These results are consistent with the task's demand. Specifically, the drivable area is large and continuous, which benefits more from high-level features, such as F5. While the lane line is narrow and fine-grained, it relies more on low-level detailed features, such as S3 and S4. These results demonstrate that our proposed adaptive segmentation decoder can automatically learn and assign appropriate weights to the different-scale feature maps for each task during the training stage.

\subsection{Ablation studies}
To evaluate our proposed module, we provide three ablation studies. First of all, we assess the effectiveness of MTL and GCA through quantitative analysis. Second, we have a gradient similarity analysis to prove that our GCA could alleviate the gradient conflicts across tasks. Finally, we have an ablation study to show the effect of the mask threshold for segmentation tasks.

\subsubsection{MTL and GCA affect}

\begin{table}[htbp]
  \centering
  \caption{Ablation study on MTL and GCA.}
  \label{tab:module_abaltion}
  \resizebox{\columnwidth}{!}{
  \begin{tabular}{@{} l c c c c c @{}}
    \toprule
    Methods                & Recall (\%) & mAP50 (\%) & mIoU (\%) & IoU (\%) & ACC (\%) \\
    \midrule
    Object only            & 92.1        & 77.5       & –         & –        & –        \\
    Drivable area only     & –           & –          & 91.0      & –        & –        \\
    Lane line only         & –           & –          & –         & 53.2     & 85.3     \\
    Segmentation only      & –           & –          & 91.3      & 53.3     & 85.4     \\
    vanilla MTL                    & 92.4        & 76.9       & 91.0      & 52.4     & 83.6     \\
    MTL with GCA (RMT-PPAD)           & 92.1        & 78.3       & 91.3      & 52.7     & 84.1     \\
    \bottomrule
  \end{tabular}
  }
\end{table}

To assess the effectiveness of MTL and the proposed GCA module, we compare the results in different configurations, as shown in Table~\ref{tab:module_abaltion}. Specifically, we train and evaluate on individual tasks first: object detection only, drivable area segmentation only, and lane line segmentation only. These are considered the baselines. In addition, we include a ``segmentation only'' setting, which jointly trains the two segmentation tasks without the detection task. This configuration is used to examine whether the two segmentation tasks negatively affect each other or can benefit from shared learning. Then, we train a vanilla MTL model for all three tasks. Finally, we integrate the GCA module into the vanilla MTL model to assess whether it alleviates negative transfer among tasks. 

The results demonstrate that the ``segmentation only'' setting improves the drivable area segmentation mIoU by 0.3\% and both IoU and ACC in lane line segmentation by 0.1\%, compared to single-task segmentation baselines. This confirms that the two segmentation tasks do not interfere with each other and may even benefit from multi-task learning. 

However, in the ``vanilla MTL'' setting, although vanilla MTL enables the model to perform all tasks simultaneously, it suffers from negative transfer, resulting in performance degradation compared to single-task baselines. Specifically, in object detection, the Recall improves slightly by 0.3\%, but mAP50 decreases by 0.6\%. In drivable area segmentation, mIoU shows no change. In lane line segmentation, both IoU and ACC decrease by 0.8\% and 1.7\%, respectively. These results indicate vanilla MTL meet negative transfer in these panoptic driving perception tasks. Both object detection and lane line segmentation performance are significantly hurt. In YOLOP \cite{wu2022yolop}, they show similar trends as well. 

To address this issue, we incorporate our proposed GCA module into the vanilla MTL model. After that, compared to vanilla MTL, mAP50 in object detection increases by 1.4\%. mIoU in drivable area segmentation improves by 0.3\%. For lane line segmentation, IoU increases by 0.3\% and ACC rises by 0.5\%. Notably, the model even outperforms the single-task baselines in mAP50 and mIoU. These results demonstrate that GCA not only effectively alleviate negative transfer by reducing task conflicts but also fosters synergy across tasks.

\subsubsection{Gradient similarity analysis}

\begin{figure*}[ht]
    \centering

    \begin{subfigure}{0.48\textwidth}
        \centering
        vanilla MTL
    \end{subfigure}%
    \hspace{0.5cm}
    \begin{subfigure}{0.48\textwidth}
        \centering
        MTL with GCA (RMT-PPAD)
    \end{subfigure}%

    \medskip

    \begin{subfigure}{0.48\textwidth}
        \centering
        \includegraphics[width=\linewidth]{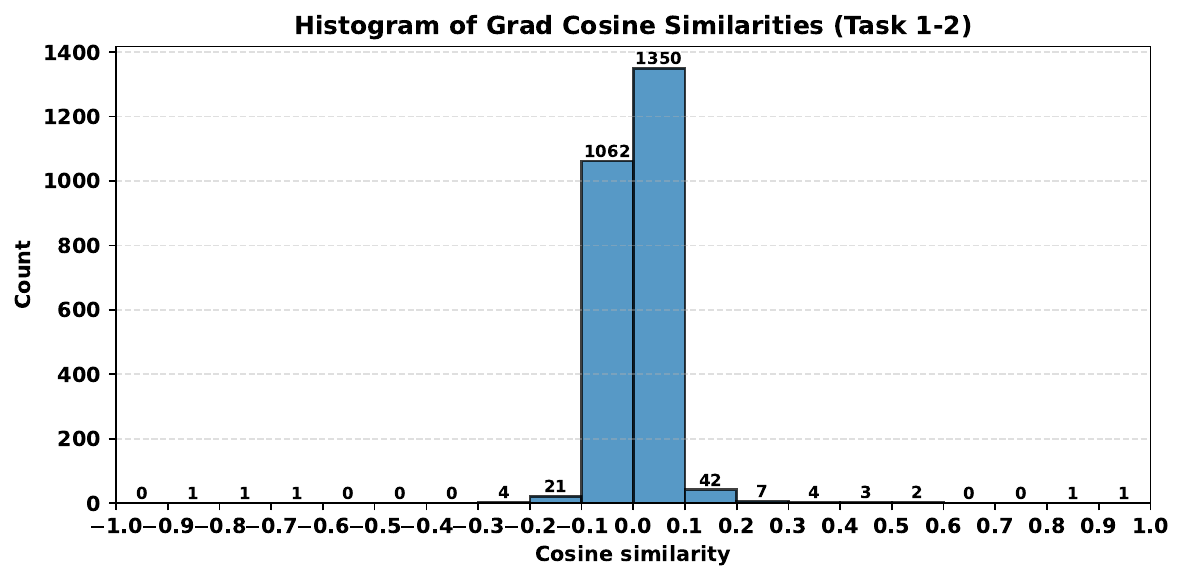}
    \end{subfigure}%
    \hspace{0.5cm}
    \begin{subfigure}{0.48\textwidth}
        \centering
        \includegraphics[width=\linewidth]{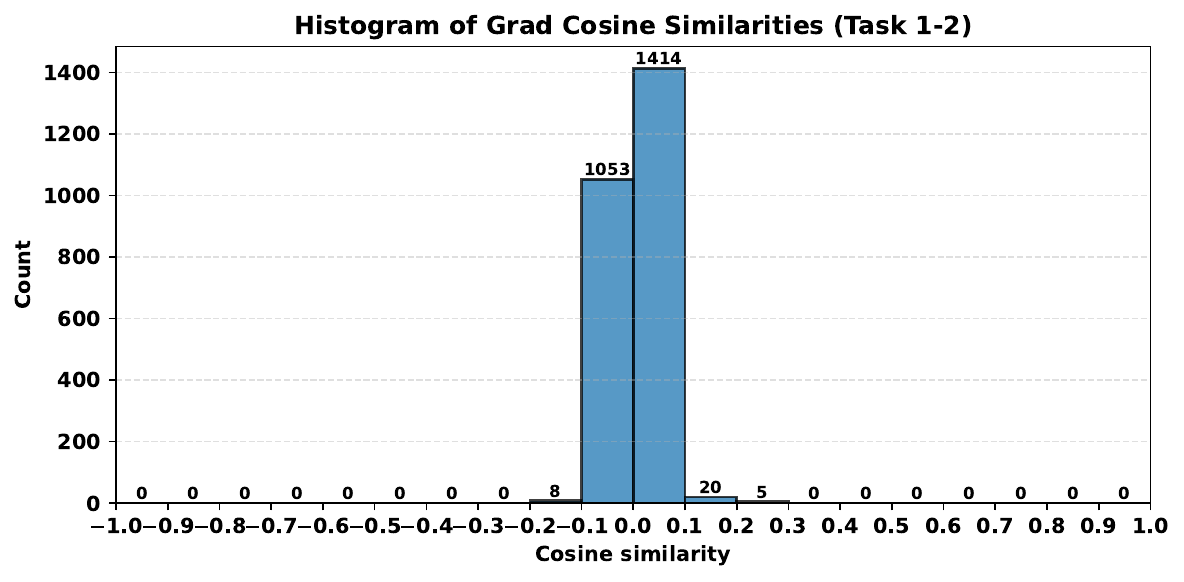}
    \end{subfigure}%

    \medskip

    \begin{subfigure}{0.48\textwidth}
        \centering
        \includegraphics[width=\linewidth]{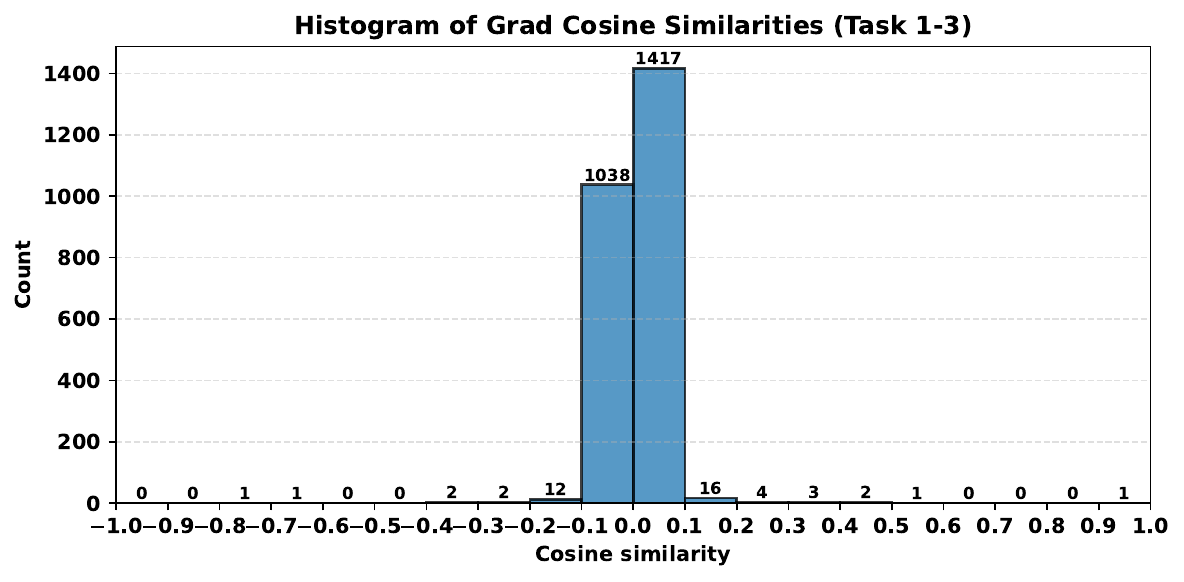}
    \end{subfigure}%
    \hspace{0.5cm}
    \begin{subfigure}{0.48\textwidth}
        \centering
        \includegraphics[width=\linewidth]{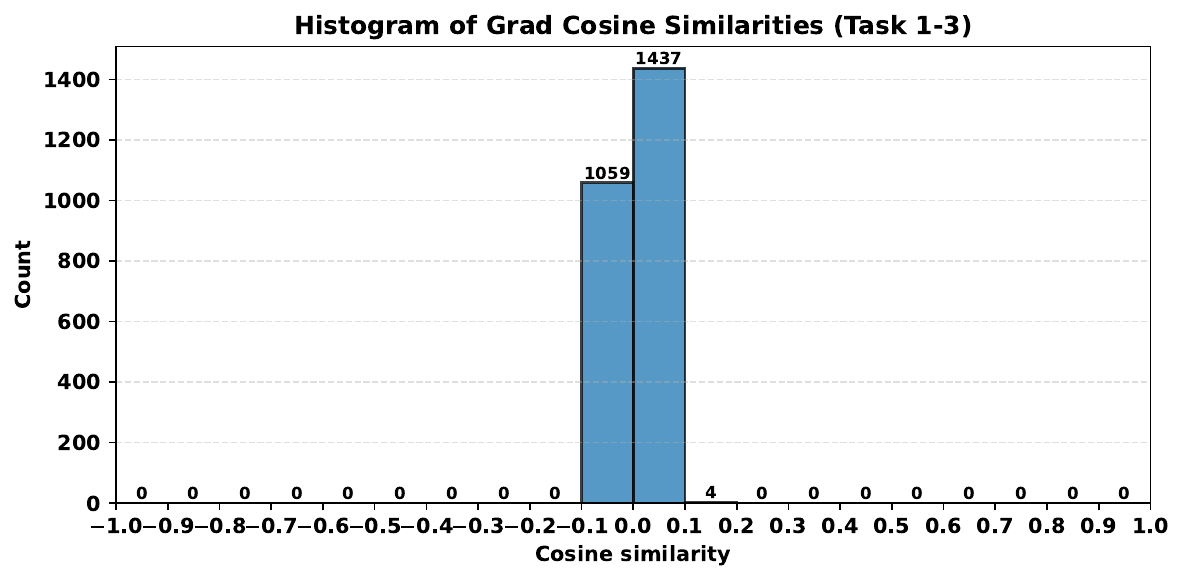}
    \end{subfigure}%

    \medskip

    \begin{subfigure}{0.48\textwidth}
        \centering
        \includegraphics[width=\linewidth]{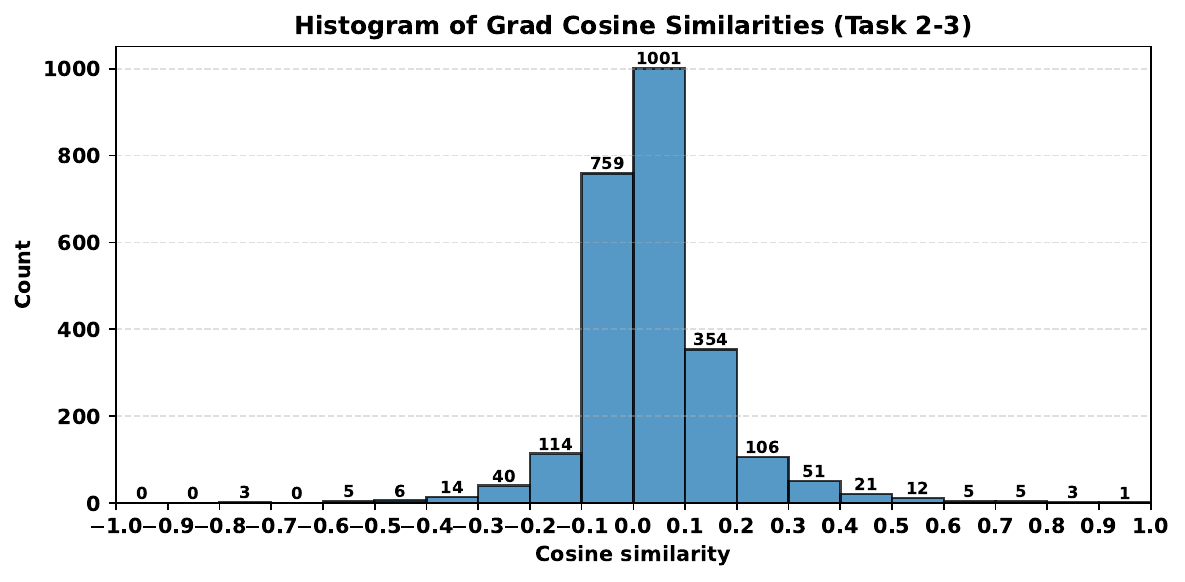}
    \end{subfigure}%
    \hspace{0.5cm}
    \begin{subfigure}{0.48\textwidth}
        \centering
        \includegraphics[width=\linewidth]{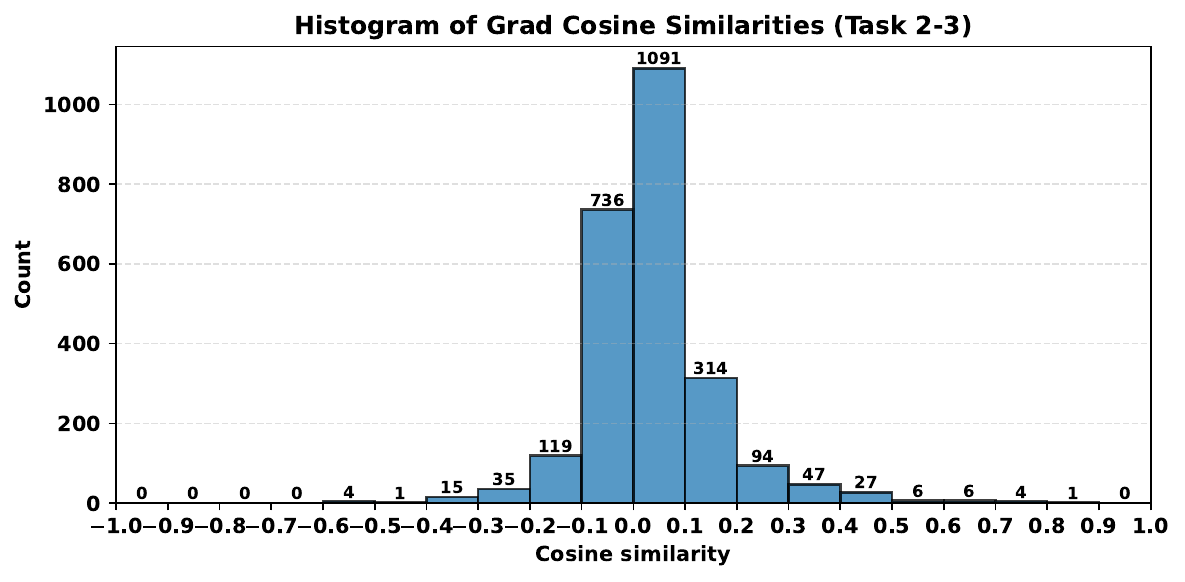}
    \end{subfigure}%

    \caption{The gradients cosine similarity of each task in vanilla MTL and RMT-PPAD. Among them, tasks 1 to 3 represent object detection, drivable area segmentation, and lane line segmentation, respectively. }
    \label{fig:cosine_similarity}
\end{figure*}

\begin{figure*}[!ht]
  \centering
  \begin{tabular}{l @{\quad} cccc}
    \rotatebox{90}{\textbf{YOLOPX}} &
      \includegraphics[width=0.22\linewidth]{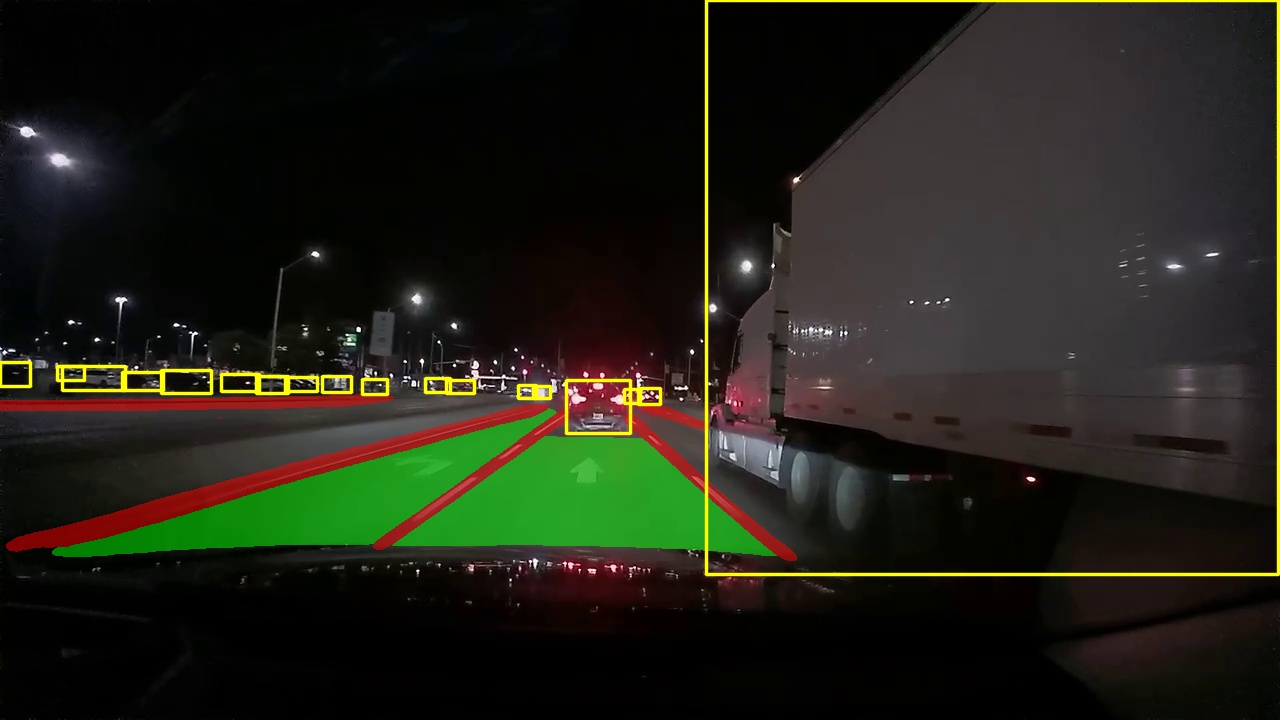} &
      \includegraphics[width=0.22\linewidth]{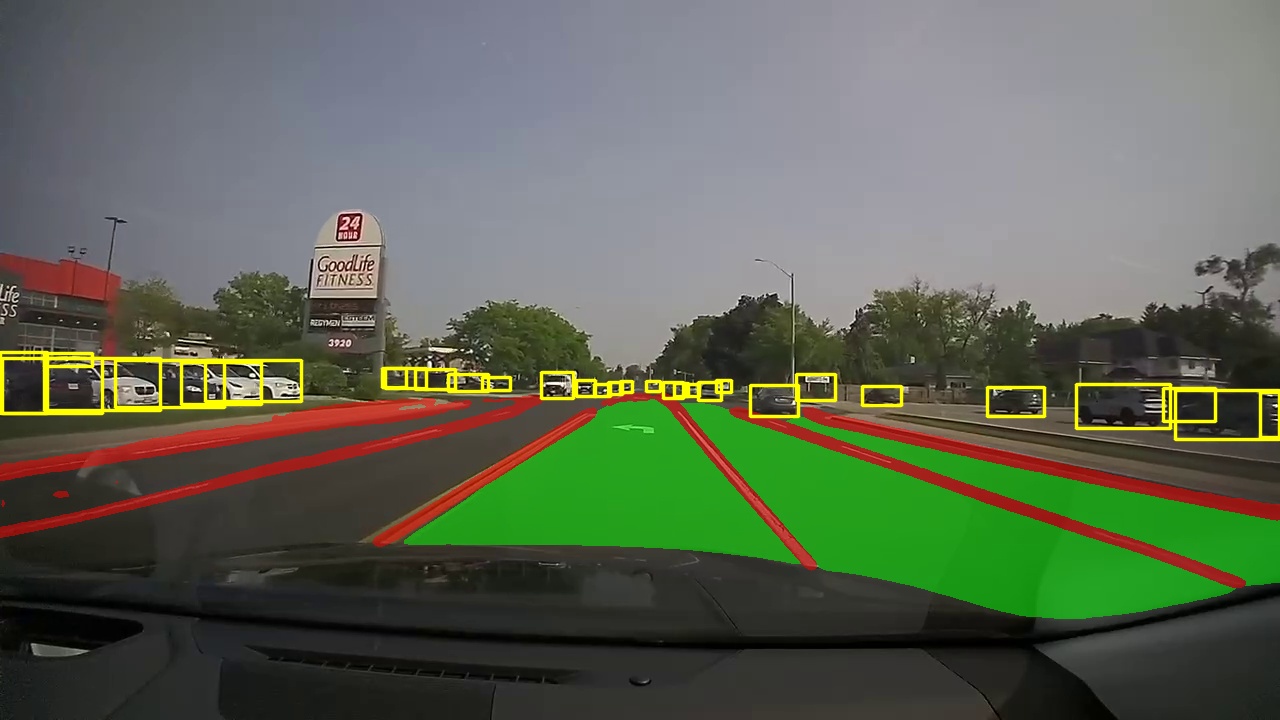} &
      \includegraphics[width=0.22\linewidth]{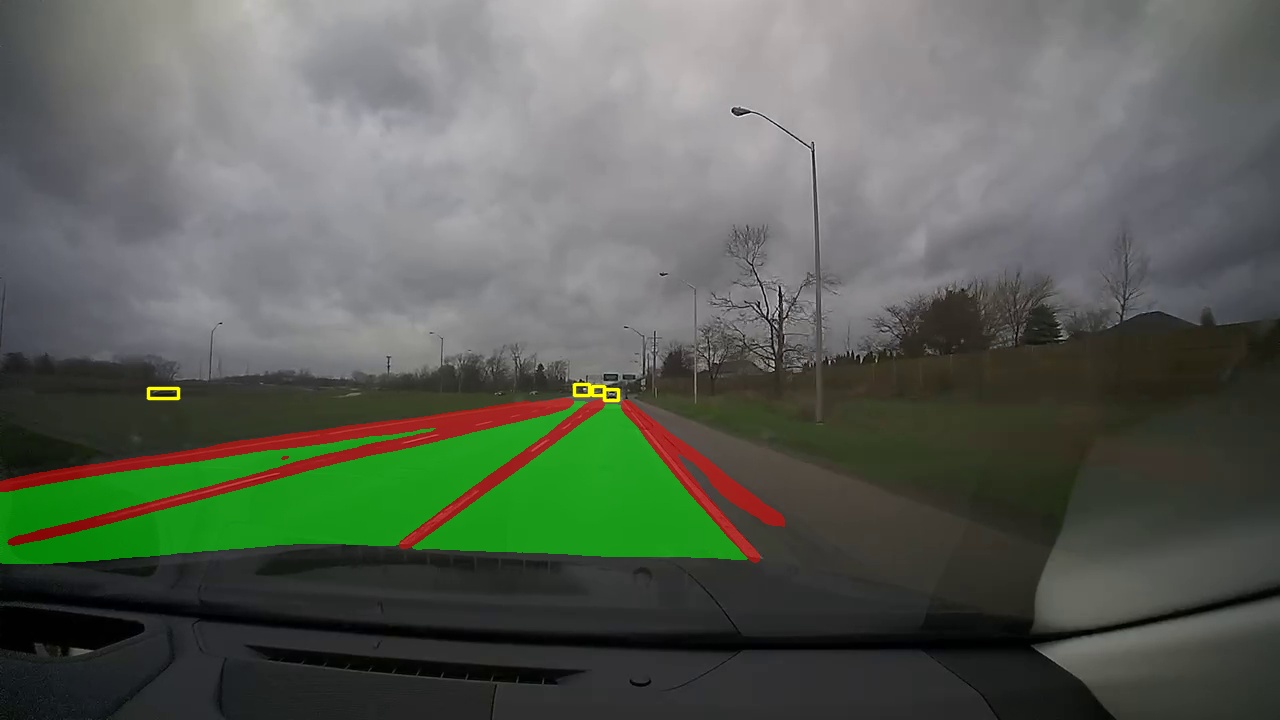} &
      \includegraphics[width=0.22\linewidth]{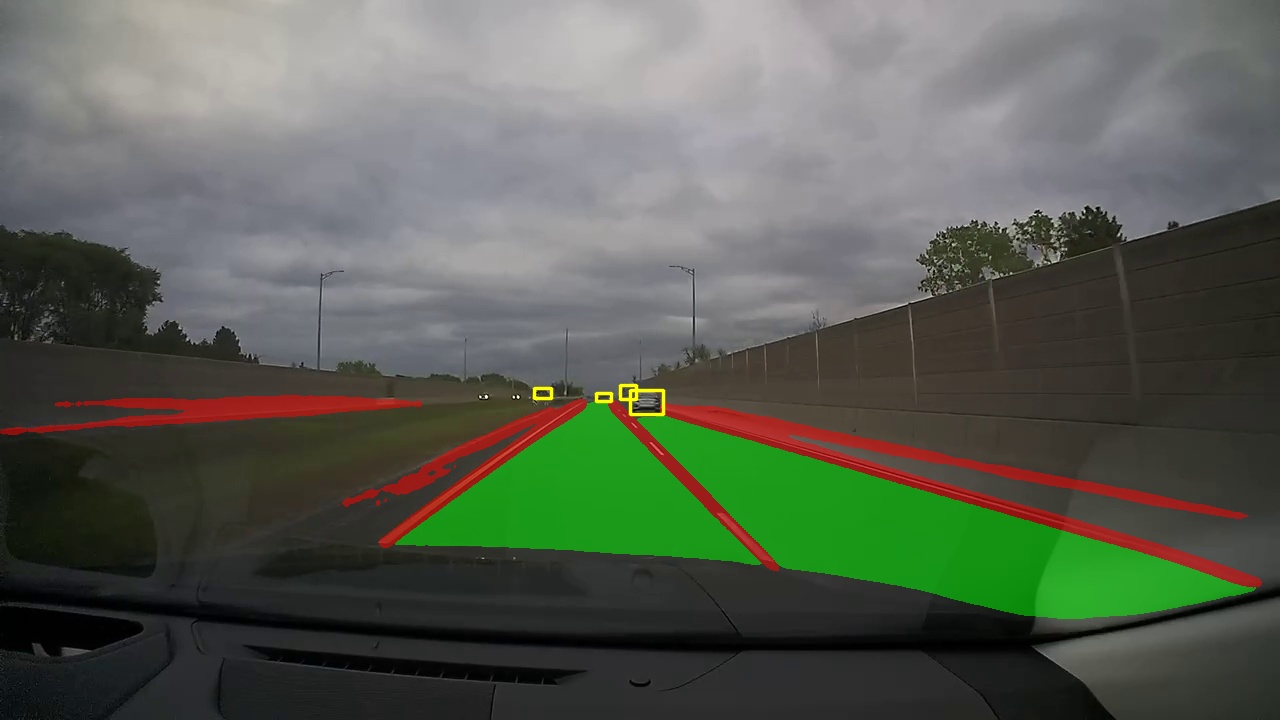} \\[6pt]

    \rotatebox{90}{\textbf{RMT-PPAD}} &
      \includegraphics[width=0.22\linewidth]{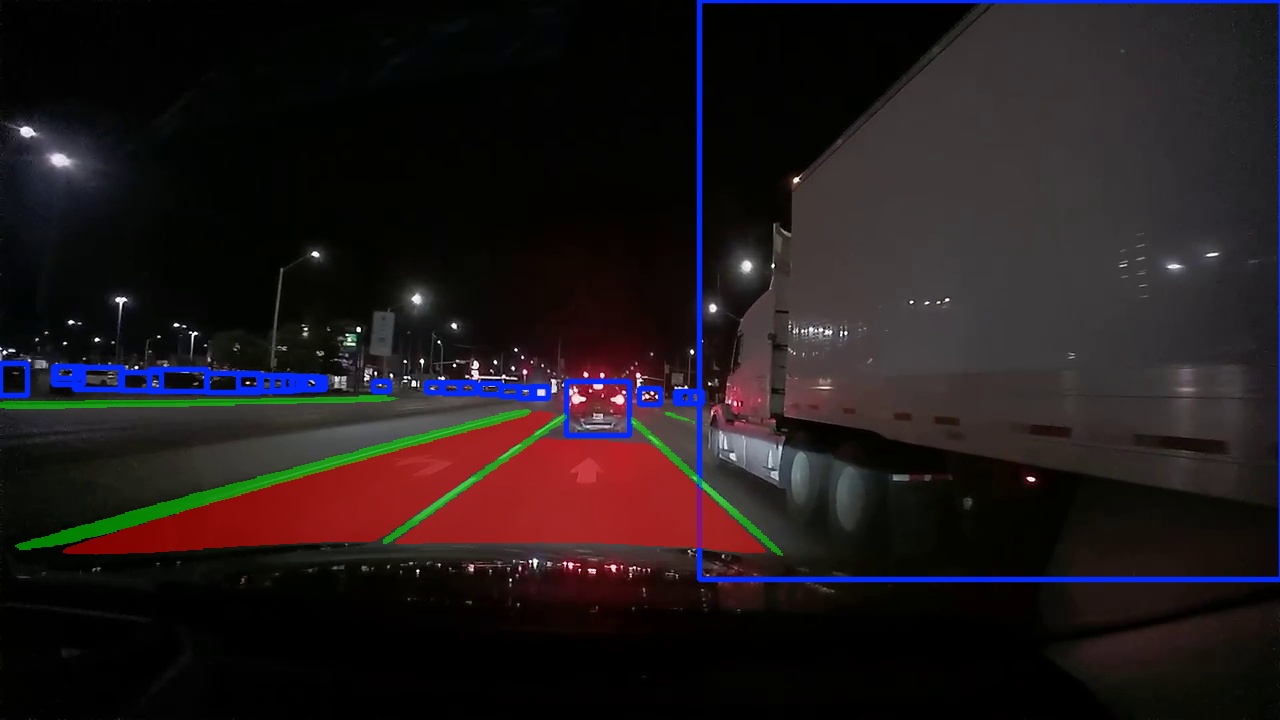} &
      \includegraphics[width=0.22\linewidth]{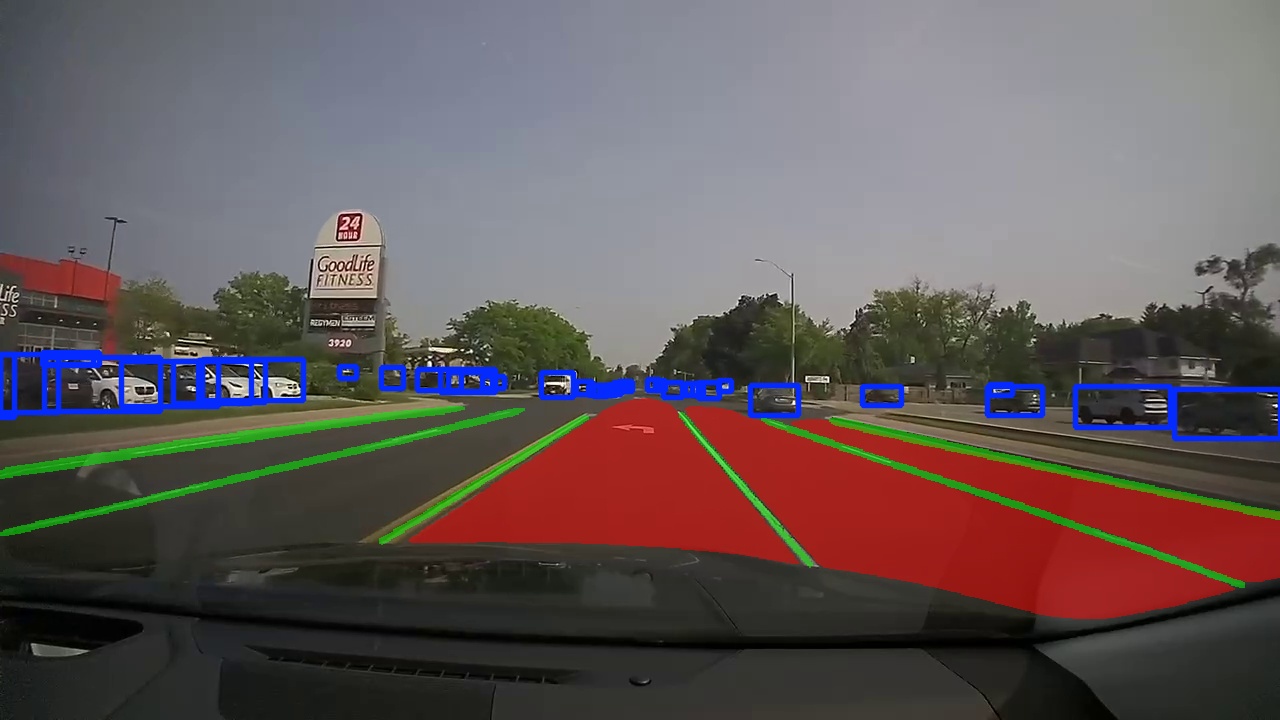} &
      \includegraphics[width=0.22\linewidth]{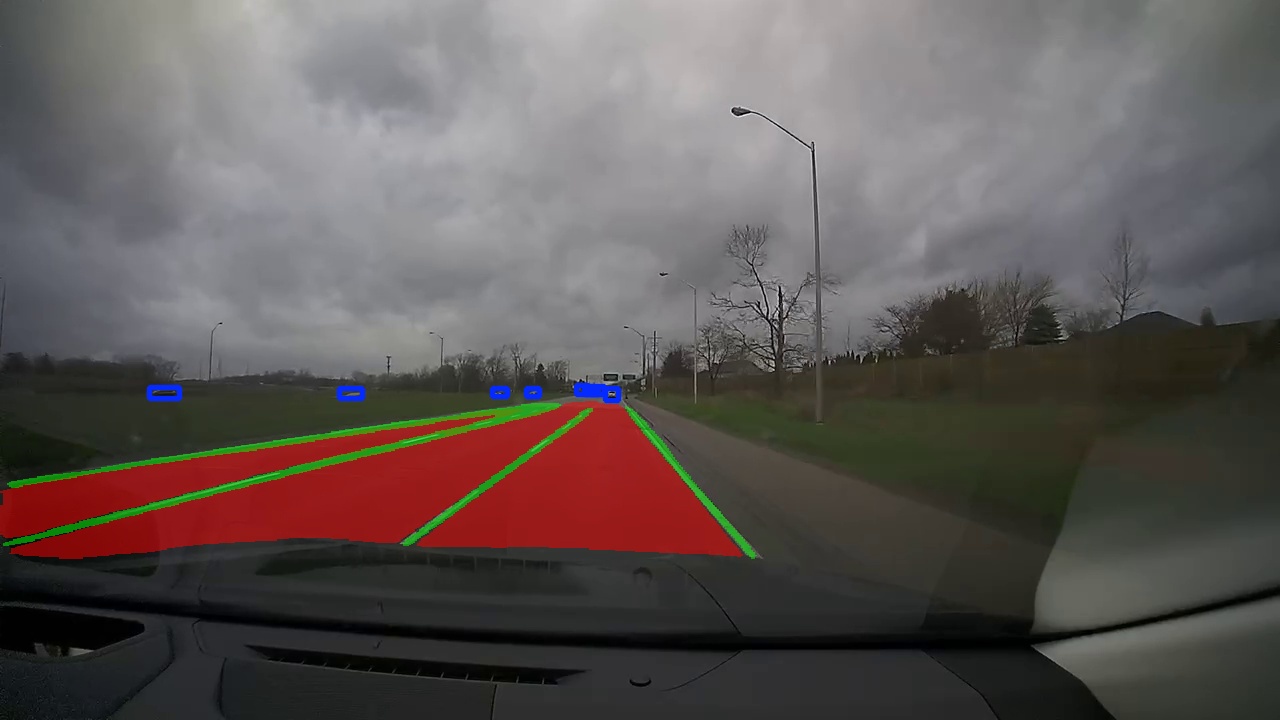} &
      \includegraphics[width=0.22\linewidth]{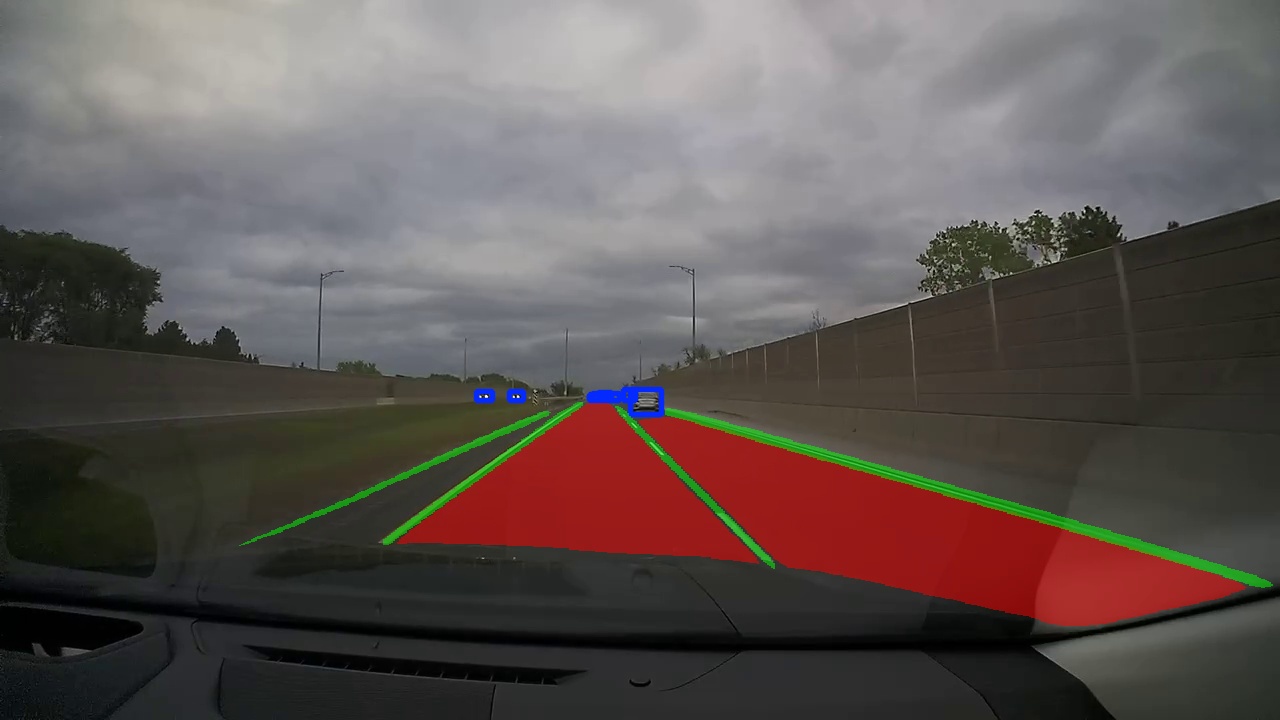} \\
  \end{tabular}
  \caption{Real road results. From left to right are nighttime, daytime, cloudy, and highway scenarios. }
  \label{fig:real_road}
\end{figure*}

To explore how GCA alleviates negative transfer in MTL, we analyze the gradient cosine similarities between each task. Fig. \ref{fig:cosine_similarity} shows the histograms of gradient cosine similarities for three task pairs in both the vanilla MTL and the MTL with GCA (RMT-PPAD) settings.

In the vanilla MTL setting, we observe that a significant part of the gradient falls in negative cosine similarities. This indicates conflicting gradient directions between tasks. This kind of conflict in gradient directions leads to inconsistent optimization across tasks. After inserting GCA, the distribution of gradient cosine similarities significantly changed. The amount of negative similarities in all task pairs is decreased. Moreover, the distributions are closer to zero. This indicates that gradients from different tasks are less conflicting and towards neutral or slightly positive directions. The gradient cosine similarities results align with the quantitative results in Table~\ref{tab:module_abaltion}, confirming that GCA effectively alleviate negative transfer across tasks.

\subsubsection{Mask threshold of segmentation tasks}
\begin{table}[ht]
  \centering
  \caption{Segmentation performance at different confidence thresholds on toy and BDD100K. mIoU for drivable area segmentation. IoU and ACC for lane line segmentation. }
  \label{tab:threshold_ablation}
  \renewcommand{\arraystretch}{1.2}
  \begin{tabularx}{\columnwidth}{@{} c *{6}{>{\centering\arraybackslash}X} @{}}
    \toprule
    \multirow{3}{*}{\textbf{Threshold}} 
      & \multicolumn{3}{c}{\textbf{Toy}} 
      & \multicolumn{3}{c}{\textbf{BDD100K}} \\
    \cmidrule(lr){2-4} \cmidrule(lr){5-7}
      & \textbf{mIoU (\%)} & \textbf{IoU (\%)} & \textbf{ACC (\%)} 
      & \textbf{mIoU (\%)} & \textbf{IoU (\%)} & \textbf{ACC (\%)} \\
    \midrule
    0.40 & \textbf{91.3} & 48.8 & \textbf{88.9} & \textbf{92.6} & 53.7 & \textbf{89.4} \\
    0.45 & \textbf{91.3} & 49.2 & 88.7 & \textbf{92.6} & 54.0 & 89.1 \\
    0.50 & 91.1 & 49.6 & 88.4 & 92.4 & 54.3 & 88.9 \\
    0.55 & 90.9 & 50.0 & 88.2 & 92.1 & 54.6 & 88.7 \\
    0.60 & 90.4 & 50.3 & 87.9 & 91.7 & 55.0 & 88.4 \\
    0.65 & 89.8 & 50.6 & 87.5 & 91.0 & 55.2 & 88.1 \\
    0.70 & 89.0 & 51.0 & 87.2 & 90.3 & 55.5 & 87.7 \\
    0.75 & 88.1 & 51.4 & 86.7 & 89.5 & 55.9 & 87.3 \\
    0.80 & 87.1 & 51.8 & 86.2 & 88.5 & 56.3 & 86.8 \\
    0.85 & 85.9 & 52.3 & 85.4 & 87.4 & 56.6 & 86.0 \\
    0.90 & 84.2 & \textbf{52.7} & 84.1 & 85.9 & \textbf{56.8} & 84.7 \\
    0.95 & 80.9 & 52.1 & 81.0 & 83.4 & 55.8 & 81.5 \\
    \bottomrule
  \end{tabularx}
\end{table}

Table \ref{tab:threshold_ablation} presents the segmentation results with different mask thresholds on both the toy and BDD100K datasets. For the drivable area segmentation, the best mIoU is achieved at a threshold of 0.40 or 0.45 on both datasets. This indicates that these thresholds are optimal for drivable area segmentation. For the lane line segmentation, since it is more sensitive to false positives, we primarily consider the IoU metric and use ACC as an auxiliary reference. The results indicate that the best IoU is at a threshold of 0.90 on both datasets. Additionally, the trends of performance change with different thresholds are the same in both toy and BDD100K datasets. This demonstrates that the optimal thresholds are determined by the task characteristic rather than the dataset. 

\subsection{Real Roads Experiments}
We also use a real-world dataset to evaluate the model's performance. Specifically, we captured several videos with a dash camera in Windsor, Ontario, Canada. We convert the videos into images frame by frame and combine them into a dataset. We compare the performance of RMT-PPAD and the next best method, YOLOPX, on this dataset. This dataset includes four scenarios: snow, cloudy, and daytime. Fig. \ref{fig:real_road} shows the results on the real road dataset. The results demonstrate that RMT-PPAD consistently maintains relatively stable performance. Compared to YOLOPX, RMT-PPAD produces smoother lane line segmentation and detects vehicles more accurately. Specifically, RMT-PPAD performs better in detecting distant and partially occluded vehicles. At nighttime, YOLOPX fails to detect vehicles at gas stations and vehicles waiting on the right side. Under cloudy and highway scenarios, YOLOPX also fails to detect distant vehicles. However, RMT-PPAD could accurately detect them in the same scenarios. These capabilities are crucial for panoptic driving perception, especially on highways, as early detection of such vehicles provides the system with more time to make decisions and control. These observations are consistent with the results shown in Fig. \ref{fig:vision_result}.

\section{CONCLUSION}
\label{sec: CONCLUSION}
In this work, we propose RMT-PPAD, a real-time transformer-based multi-task model with a GCA module. RMT-PPAD jointly addresses object detection, drivable area segmentation, and lane line segmentation in a single model while maintaining real-time inference. Specifically, we introduce a lightweight GCA module, which effectively alleviates negative transfer by adaptively fusing shared and task-specific features. Additionally, to avoid manual design for different segmentation heads based on prior knowledge, we propose the adaptive segmentation decoder, which learns task-specific multi-scale feature weights automatically during the training stage. Moreover, we identified and resolved the lane line label inconsistency issue, improving the fairness of lane line segmentation evaluation. Compared to open-source MTL models on the BDD100K dataset, the results demonstrate that RMT-PPAD achieves SOTA performance while remaining real-time inference. Ablation studies and gradient analyses further validate the contributions of each module. Moreover, we test RMT-PPAD and YOLOPX on real-world scenarios. The results show that RMT-PPAD consistently delivers stable performance. In future work, we plan to investigate model compression techniques to reduce the parameters of the MTL model while maintaining its strong performance. 

\bibliographystyle{IEEEtran}

\bibliography{ref}

\end{document}